\documentclass[letterpaper, 10 pt, conference]{ieeeconf}  % Comment this line out if you need a4paper
\IEEEoverridecommandlockouts                              % This command is only needed if 
\makeatletter
\let\NAT@parse\undefined
\makeatother

\usepackage[numbers,sort]{natbib}
\usepackage{times}
\usepackage{graphicx}
\usepackage{bbm}
\usepackage{amsmath}
\usepackage{amssymb}
\usepackage{color, colortbl}
\usepackage{tabularx}
\usepackage{hyperref}
\usepackage{algorithm}
\usepackage[noend]{algorithmic}
\usepackage[normalem]{ulem}

\usepackage{multirow}

\usepackage{etoolbox}
\makeatletter
\patchcmd{\@makecaption}
  {\scshape}
  {}
  {}
  {}
\makeatletter
\patchcmd{\@makecaption}
  {\\}
  {.\ }
  {}
  {}
\makeatother

\definecolor{LightCyan}{rgb}{0.84,0.94,0.95}
\definecolor{Fail}{rgb}{0.84,0.84,0.84}

\selectfont
\title{\LARGE \bf
DexPilot: Vision Based Teleoperation of \\ Dexterous Robotic Hand-Arm System}
\author{ Ankur Handa$^{* \dagger}$ \qquad Karl Van Wyk$^{* \dagger}$ \qquad Wei Yang$^{\dagger}$ \qquad Jacky Liang$^{\ddagger}$ \qquad Yu-Wei Chao$^{\dagger}$ \\
Qian Wan$^{\dagger}$ \qquad Stan Birchfield$^{\dagger}$\qquad Nathan Ratliff$^{\dagger}$\qquad Dieter Fox$^{\dagger}$
\thanks{$^{*}$ Equal Contribution}
\thanks{$^{\dagger}$NVIDIA, USA}%
\thanks{$^{\ddagger}$CMU, Pittsburgh, PA, USA}
\thanks{$^@${\tt\small \{ahanda,kvanwyk,dieterf\}@nvidia.com}
}}

\begin{document}

\makeatletter
\let\@oldmaketitle\@maketitle% Store \@maketitle
\renewcommand{\@maketitle}{\@oldmaketitle% Update \@maketitle to insert...
  \includegraphics[width=\linewidth]
    {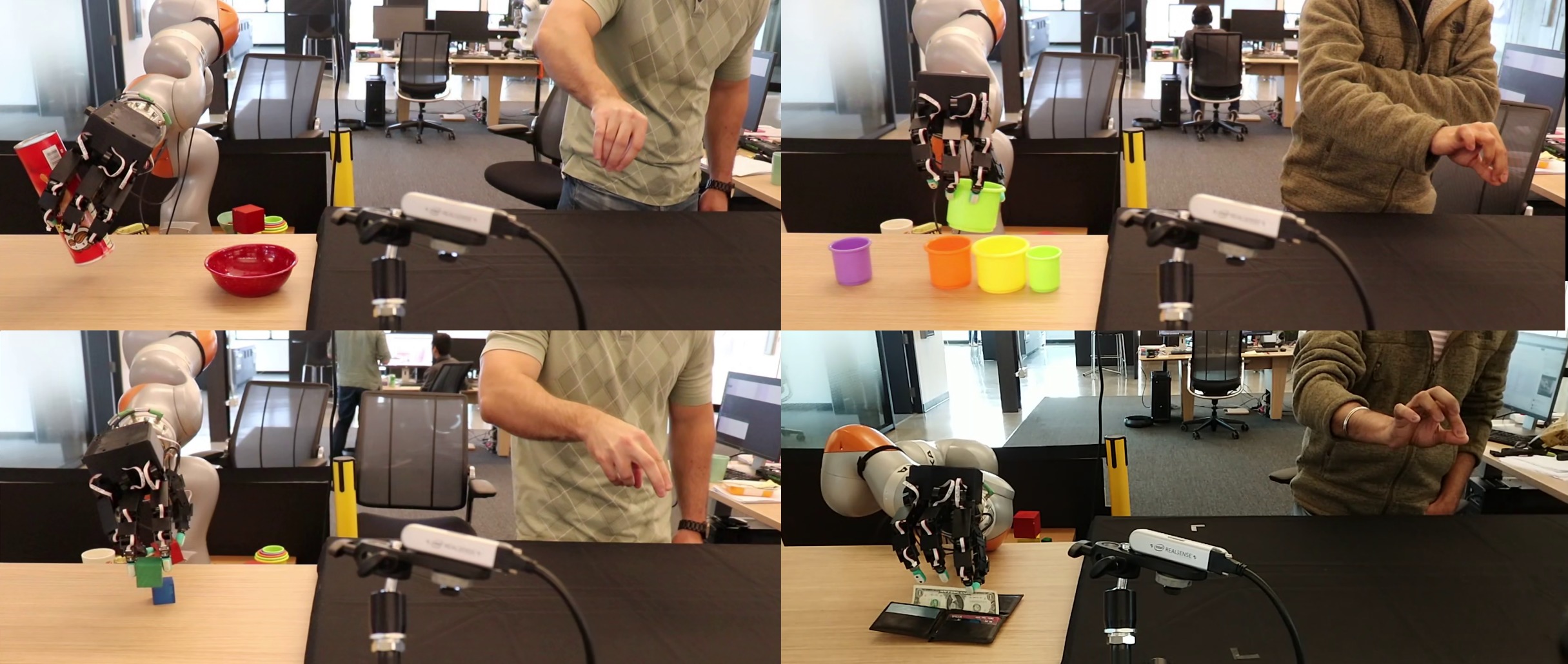} \\[0.25em]
  \refstepcounter{figure}\footnotesize{Fig.~\thefigure. DexPilot enabled teleoperation across a wide variety of tasks, \textit{e.g.}, rectifying a Pringles can and placing it inside the red bowl (upper-left), inserting cups (upper-right), concurrently picking two cubes with four fingers (lower-left), and extracting money from a wallet (lower-right). Videos are available at \url{https://sites.google.com/view/dex-pilot}.}
  \label{fig:title} \medskip \vspace{-10pt}}% ... an image
\makeatother

% \makeatletter
% \let\@oldmaketitle\@maketitle% Store \@maketitle
% \renewcommand{\@maketitle}{\@oldmaketitle% Update \@maketitle to insert...
%   \includegraphics[width=0.5\linewidth]
%     {figs/collage_demos.png} 
%  \includegraphics[width=0.5\linewidth]
%     {figs/studio_2.jpg} \\[0.25em]
%   \refstepcounter{figure}\footnotesize{Fig.~\thefigure. Tele-op studio. One of these images should capture the hand-arm system.}
%   \label{fig:title} \medskip \vspace{-10pt}}% ... an image
% \makeatother

\maketitle
\thispagestyle{empty}
\pagestyle{empty}

%%%%%%%%%%%%%%%%%%%%%%%%%%%%%%%%%%%%%%%%%%%%%%%%%%%%%%%%%%%%%%%%%%%%%%%%%%%%%%%%

\begin{abstract}
Teleoperation offers the possibility of imparting robotic systems with sophisticated reasoning skills, intuition, and creativity to perform tasks. 
However, current teleoperation solutions for high degree-of-actuation (DoA), multi-fingered robots are generally cost-prohibitive, while low-cost offerings usually provide reduced degrees of control. Herein, a low-cost, vision based teleoperation system, DexPilot, was developed that allows for complete control over the full 23 DoA robotic system by merely observing the bare human hand. 
DexPilot enables operators to carry out a variety of complex manipulation tasks that go beyond simple pick-and-place operations. This allows for collection of high dimensional, multi-modality, state-action data that can be leveraged in the future to learn sensorimotor policies for challenging manipulation tasks. The system performance was measured through speed and reliability metrics across two human demonstrators on a variety of tasks. The videos of the experiments can be found at \url{https://sites.google.com/view/dex-pilot}.
\end{abstract}

%%%%%%%%%%%%%%%%%%%%%%%%%%%%%%%%%%%%%%%%%%%%%%%%%%%%%%%%%%%%%%%%%%%%%%%%%%%%%%%
\section{Introduction}
%The main question this research seeks to answer is the following: can a low-cost, glove-free tele-operation system be constructed that leverages innovations in machine vision, optimization, motion generation, and GPU compute to drive a multi-fingered, highly actuated robot system to solve a wide variety of grasping and manipulation tasks? 
%The developed DexPilot is such a system that operates only on multi-camera RGB-D observations of the bare human hand. Although this approach does not relay tactile feedback to the user, DexPilot still enables successful tele-operation due to the sophisticated human visuomotor control \cite{johansson2009coding}, spatial reasoning skill, and creativity. Therefore, out-of-body tele-operation is still successful despite tactile sensory deprivation. 

Robotic teleoperation has been researched for decades with applications in search and rescue \cite{norton2017analysis}, space \cite{diftler2012robonaut}, medicine \cite{sterbis2008transcontinental}, prosthetics \cite{zhuang2019shared}, and applied machine learning \cite{zhang2018deep}. The primary motivation of teleoperation is to allow a robot system to perform complex tasks by harnessing the cognition, creativity, and reactivity of humans through a human-machine interface (HMI). Through the years, many advancements have been made in this research field including the incorporation of haptic feedback \cite{HaptX, CyberGlove} and improved human skeletal and finger tracking \cite{Stenger:etal:PAMI2006,Tompson:etal:ToG2014,Yuan:etal:CVPR2017,Ge:etal:ECCV2018}. Nevertheless, the cost of many high DoA teleoperation or tracking systems is prohibitive. This research bridges this gap by providing a low-cost, markerless, glove-free teleoperation solution that leverages innovations in machine vision, optimization, motion generation, and GPU compute. Despite its low-cost, the system retains the ability to capture and relay fine dexterous manipulation to drive a highly actuated robot system to solve a wide variety of grasping and manipulation tasks. Altogether, four Intel RealSense depth cameras and two NVIDIA GPUs in combination with deep learning and nonlinear optimization produced a minimal-footprint, dexterous teleoperation system. Despite the lack of tactile feedback, the system is highly capable and effective through human cognition. This result corroborates human gaze studies that indicate that humans learn to leverage vision for planning, control, and state prediction of hand actions \cite{johansson2009coding} prior to accurate hand control. Therefore, the teleoperation system exploits the human ability to plan, move, and predict the consequence of their physical actions from vision alone; a sufficient condition for solving a variety of tasks. The main contributions are summarised below:

\begin{itemize}
    \item Markerless, glove-free and entirely vision-based teleoperation system that dexterously articulates a highly-actuated robotic hand-arm system with direct imitation.
    \item Novel cost function and projection schemes for kinematically retargeting human hand joints to Allegro hand joints that preserve hand dexterity and feasibility of precision grasps in the presence of hand joint tracking error.
    \item Demonstration of teleoperation system on a wide variety of tasks particularly involving fine manipulations and dexterity, \textit{e.g.}, pulling out paper currency from wallet and grasping two cubes with four fingers as shown in Fig. \ref{fig:title}. 
    \item System assessment across two trained human demonstrators --- also called pilots --- revealed that high task success rates can be achieved despite the lack of tactile feedback.
\end{itemize}

\section{Related Work}
Teleoperation via the human hand is traditionally done using either pure vision or sensorized glove-based solutions and relies on accurate hand tracking. Early vision-based solutions to hand tracknig include Dorner \textit{et al.} \cite{Dorner:etal:MS1994} who used colour markers on the hand to track the 2D positions of joints and fit a skeleton model to obtain the hand pose and finger joints in 3D. Similarly, Theobalt \textit{et al.} , \cite{Theobalt:etal:SIGGRAPH2004} use colour markers to track the motion of the hand in high speed scenarios in sports. Wang \textit{et al. }\cite{Wang:etal:2009} propose a fully coloured glove to track the joints and pose of the hand for augmented reality (AR) and virtual reality (VR) applications. Impressive strides have also been made in the past few years in bare human hand tracking particularly with deep learning \cite{Tompson:etal:ToG2014,Oberweger:etal:2015, Yuan:etal:CVPR2017,Ge:etal:ECCV2018,Cao:etal:arXiv2018, Oberweger:etal:PAMI2019}. Glove- and marker-free hand tracking is attractive as it produces minimal external interference with the hand and reduces the system physical footprint, but reliability and accuracy with such approaches have not yet reached desireable performance levels. Moreover, most of these works study hand tracking in isolation without evaluation of tracking performance in controlling a physical robot. Li \textit{et al.} \cite{li:icra2019} proposed vision-based teleoperation through deep learning that calculated Shadow robotic hand joint angles from observed human hand depth images. However, this approach was not extended to a full robotic system and the mapping results had performance issues as later discussed in Section \ref{sec:kinematic_retargeting}. Antotsiou \textit{et al.} \cite{Antotsiou:etal:ECCVW2018} show teleoperation of an anthropomorphic hand in simulation using depth based hand tracking. They focus on very simplistic grasping \textit{e.g.} picking a ball on a flat table and do not show the level of dexterity as seen herein. In general, hand tracking with vision remains a very challenging problem due to self-similarity and self-occlusions observed in the hand.

On the other end of the spectrum, various commercial glove- or marker-based solutions offer accurate hand tracking. Additionally, approaches from CyberGlove \cite{CyberGlove} and HaptX \cite{HaptX} do provide the user with tactile feedback in the form of physical resistance and spatial contact, but this benefit comes with heightened cost and additional bulk to the user. Sometimes these systems still require an external hand pose estimation module as in \cite{Kumar:etal:RAS2015} to enable a freely moving mobile hand as with DexPilot. Still, these approaches are important to investigate how relaying tactile feedback to human pilots improve teleoperation capability considering that numerous neurophysiological studies have uncovered the importance of tactile feedback in human sensorimotor control of the human hand \cite{johansson2009coding, johansson1984roles, johansson1983tactile}. Alternatively, motion capture systems provide accurate point tracking solutions, but can be expensive with multi-camera systems. Moreover, when tracking all the joints of the hands, the correspondence problem between markers on the fingers and cameras still needs to be solved. For instance, Han \textit{et al.} \cite{Han:etal:2018} track hands with an array of OptiTrack cameras while solving the correspondence problem via a neural network. They show impressive tracking of two hands in VR settings. Zhuang \textit{et al.} \cite{zhuang2019shared} leverage a motion capture system and myoelectric device for teleoperative control over a hand-arm system. The system worked convincingly well as a prosthetic HMI, but experiments did not reveal how well the system works for intricate in-hand manipulation tasks (as assessed herein). Overall, these solutions are viable alternatives, but are significantly more expensive and with larger physical footprints than DexPilot. Finally, some of these approaches only consider the problem of tracking hand movements, and are not complete teleoperation solutions like DexPilot.

We also note a few low-cost hardware solutions that have emerged lately. For example, \cite{Liu:etal:ICRA2019} and \cite{Liu:etal:IROS2017} showed that low-cost solutions for teleoperation with a glove can be constructed though they tend to be quite bulky and have wires tethered to them. 

%users feel the cognitive load using controllers with Oculus.
% Add Jeannette's paper on hand segmentation \cite{Kokic:etal:IROS2019} etc.

% Acquiring human demonstration:

% VR + simulation~\cite{kumar:humanoids2015,lin:siggraphasiaw2016,rajeswaran:rss2018,liu:icra2019}

% VR + teleoperation~\cite{zhang:icra2018}

%\includegraphics[width=0.5\linewidth]{figs/hand_design_v2.png} 

\section{Hardware}

\begin{figure*}[t]
  \centering
  \includegraphics[width=\linewidth]{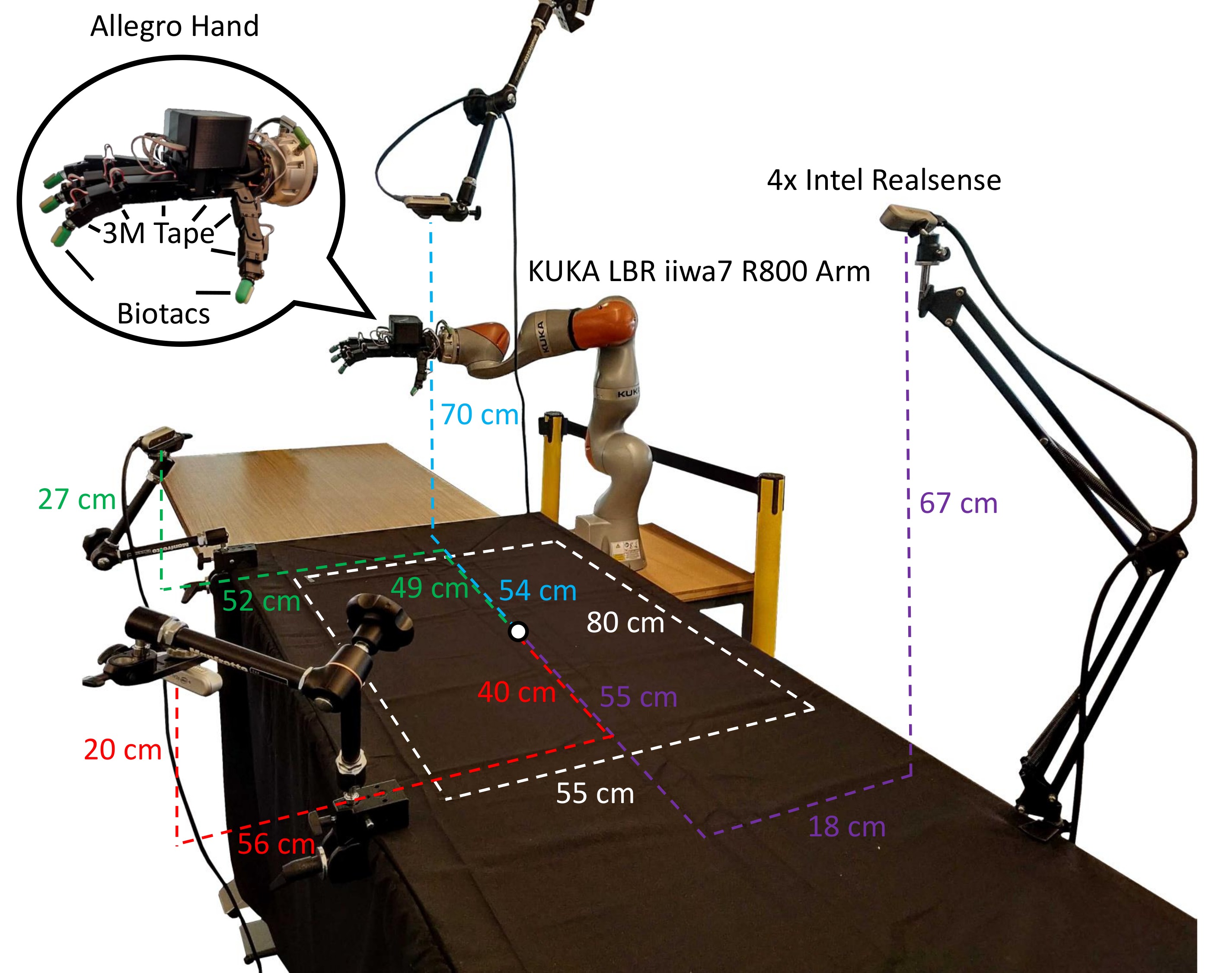}
  \vspace{-5mm}
  \caption{\footnotesize{Studio is composed of four cameras pointing towards the table over which the user moves their hand with the hand-arm system in close proximity to enable line of sight driven teleoperation.}}
  \label{fig:studio}
\end{figure*}

The teleoperation setup comprised of a robot system and an adjacent human pilot arena as shown in Fig.~\ref{fig:studio}. The robot system was a KUKA LBR iiwa7 R800 series robot arm and a Wonik Robotics Allegro hand. The Allegro hand was retrofitted with four Syntouch Biotac tactile sensors at the fingertips and 3M TB641 grip tape applied to the inner surfaces of the phalanges and palm. The rubbery surfaces of both the Biotac sensors and 3M tape augmented the frictional properties of the hand while the Biotacs themselves produced 23 signals that can later be used to learn sensorimotor control from demonstrations. In total, the robotic system has 92 tactile signals, 23 joint position signals, and 23 joint torque actions. The human arena was a black-clothed table that housed four calibrated and time-synchronized Intel Realsense D415 RGB-D cameras spatially arranged to yield good coverage of the observation volume within the ideal camera sensing range. Ideally, depth observations should remain within 1 $m$ for each camera; otherwise, the depth quality degenerates. The human arena is directly adjacent to the robot to improve line-of-sight and visual proximity since teleoperation is entirely based on the human vision and spatial reasoning. The teleoperation work volume is 80 $cm$ $\times$ 55 $cm$ $\times$ 38 $cm$.

\section{Architecture}
\begin{figure*}[t]
    \centering
    \includegraphics[width=1.0\linewidth]{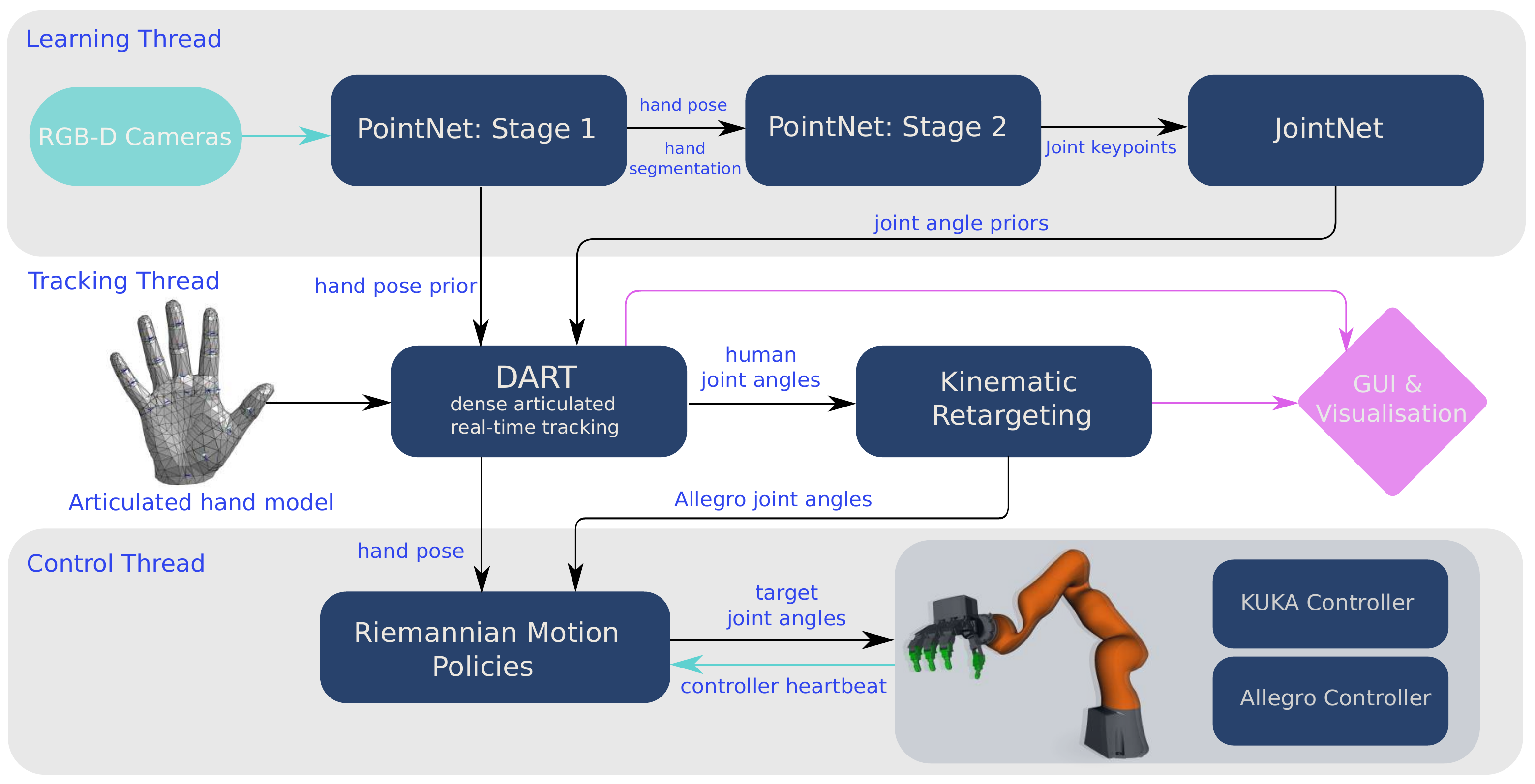}
    \vspace{-5mm}
    \caption{\footnotesize{The system is comprised of three threads that operate on three different computers. The learning thread provides hand pose and joint angle priors using fused input point cloud coming from four cameras from the studio. The tracking thread runs DART for hand tracking with the priors as well as kinematic retargeting needed to map human hand configuration to allegro hand. The control thread runs the Riemannian motion policies to provide the target joint commands to the KUKA and allegro hand given the hand pose and joint angles.}}
    \label{fig:architecture}
\end{figure*}

To produce a natural-feeling teleoperation system, an imitation-type paradigm is adopted. The bare human hand motion --- pose and finger configuration --- was constantly observed and measured by the visual perception module. The human hand motion is then relayed to the robot system in such a way that the copied motion is self-evident. This approach enables the human pilot to curl and arrange their fingers, form grasps, reorient and translate their palms, with the robot system following in a similar manner. DexPilot relies heavily on DART \cite{Schmidt:etal:RSS2014} which forms the backbone of tracking the pose and joint angles of the hand. In the following, we explain the main components of the overall system: 1) DART for hand tracking (Section \ref{section:DART}). 2) deep neural networks for human hand state estimation and robustifying DART (Section \ref{sec:vision}). 3) human hand state refinement with DART and its conversion through nonlinear optimization to Allegro hand states (Section \ref{sec:kinematic_retargeting}) 4) motion generation and control through Riemannian Motion Policies (RMPs) and torque-level impedance controllers (Section \ref{sec:RMPs}). The fully system architecture is shown in Fig. \ref{fig:architecture} where these components are daisy chained. Altogether, the system produces a latency of about one second.

% The full system architecture is shown in Fig. \ref{fig:architecture} with three main components: 1) deep neural networks for human hand state estimation (Section \ref{sec:vision}), 2) human hand state refinement with DART and its conversion through nonlinear optimization to Allegro hand states (Section \ref{sec:kinematic_retargeting}), and 3) motion generation and control through Riemannian Motion Policies (RMPs) and torque-level impedance controllers (Section \ref{sec:rmp}).

\section{DART}
\label{section:DART}
DART \cite{Schmidt:etal:RSS2014} produces continuous pose and joint angle tracking of the human hand by matching an articulated model of the hand against an input point cloud. The human hand model was obtained from  \cite{MANO:SIGGRAPHASIA:2017} and turned into a single mesh model \cite{Hasson:etal:CVPR19}. With CAD software, the fingers of the mesh model were separated into their respective proximal, medial, and distal links, and re-exported as separate meshes along with an associated XML file that described their kinematic arrangement. In total, the human hand model possessed 20 revolute joints: four joints per finger with one abduction joint and three flexion joints. 

\section{Estimating hand pose and joint angles with neural networks}
\label{sec:vision}
Since DART is a model-based tracker that relies on non-linear optimisation, it needs initialisation which typically comes from estimates from previous frame. If this initialisation is not within the basin of convergence, the tracker can fail catastrophically. Often, DART can match the hand model to the raw point cloud in regions of spurious local minima leading to tracking failures every few minutes. Therefore, to allow for tracking over long periods of time --- as needed for teleoperation --- hand pose priors and hand segmentation can be implemented to prevent the hand model from converging to incorrect local minima. One way of getting hand pose priors would to be train a neural network on a large dataset of human hand poses. Although various datasets \cite{Tompson:etal:ToG2014, Yuan:etal:CVPR2017} exist that provide annotations for hand pose as well as segmentation, they were not suitable for this setting primarily due to different sensor characteristics and lack of hand poses required for dexterous manipulation. 

\begin{figure}
  \centering
  \includegraphics[width=0.7\linewidth]{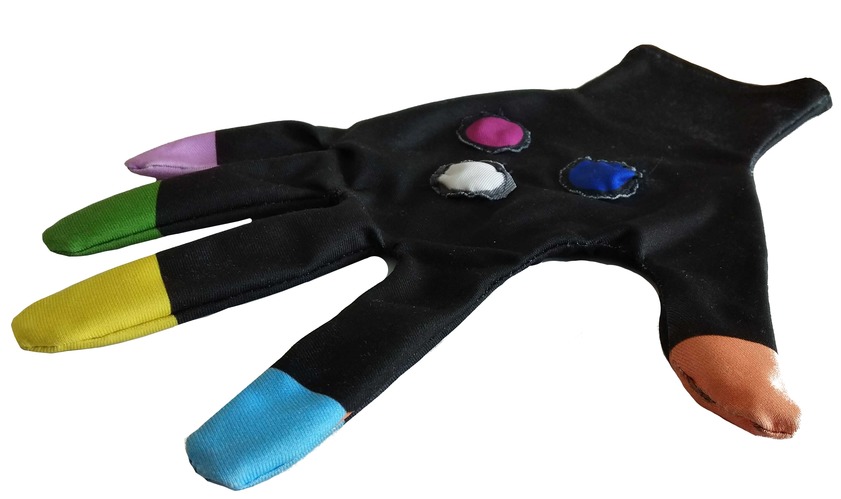}
  \vspace{-3mm}
  \caption{\footnotesize{The colour glove used in the first phase to obtain hand pose and segmentation for DART. Unique colours are printed such that annotation generation is trivial with OpenCV colour thresholding. The colours on the back of the palm can uniquely determine the pose of the hand.}}
  \label{fig:colour glove}
\end{figure}

Therefore, a fabric glove (shown in Fig.~\ref{fig:colour glove}) with coloured blobs was initially used as an effective solution for obtaining a hand pose prior with a deep neural network. The data collection proceeded in two phases. In the first phase, the user wore the glove to obtain hand pose priors for DART to track human hand robustly. This process generated hand pose and joint angle annotations for raw depth maps from the RGB-D cameras for the second phase. The second phase uses these annotations and operates on raw point cloud from corresponding depth maps and frees the user from having to wear the glove. We explain differetnt phases below.

\paragraph{\textbf{First Phase}} The color glove is inspired by \cite{Dorner:etal:MS1994, Wang:etal:2009} who used it for hand tracking. The glove has coloured blobs at the finger tips and three at the back of the palm. It is worth clarifying that hand tracking includes both the hand pose and the joint angles of the fingers. In our set-up, the user moves their hand over a table in a multi-camera studio with four Intel RealSense D415 RGB-D cameras pointing downwards to the table. The problem of hand pose estimation with glove is formulated via keypoint localisation: ResNet-50 \cite{He:etal:CVPR2016} with spatial-softmax is used to regress from an RGB image to the 2D locations of the centers of the coloured blobs on the glove. We call this network GloveNet. The coloured blobs at finger-tips are also regressed but were found to be not helpful in full hand tracking in the end and therefore we only use the predictions of the blobs on the back of the palm for hand pose estimation. We explain that in detail in the appendix. 

The hand pose can be estimated by three unique keypoints as indicated by three different coloured blobs at the back of the palm of glove. To obtain annotations for the centers of the blobs, HSV thresholding in OpenCV is used to generate segmentations and compute the centeroids of these segmented coloured blobs. To aid segmentation for high quality annotations, the user wears a black glove with coloured blobs and moves the hand over a table also covered with black cloth. The pose of the hand can be obtained via predicted 2D locations of the blobs from all four cameras: the 2D keypoints are converted to their corresponding 3D locations using the depth values resulting in each blob having four 3D predictions in total from four cameras. These 3D locations are filtered and temporally smoothed to obtain the hand pose. Hand segmentation is also obtained by removing the 3D points that fall outside the bounding volume of the hand. The dimensions of this volume were obtained heuristically from the hand pose obtained from the neural network predictions. Crucially, DART now only optimises on the segmented hand points, preventing the hand model from sliding out to points on the arm as often observed when a full point cloud is used. It is worthwhile remembering that DART does not use RGB images --- the glove only provided pose priors and aided hand segmentation --- and therefore the result of DART with hand pose priors and segmentation in the first phase is generating annotations for raw point cloud captured with the cameras for second phase which can operate on the bare human hand.

% \section{PointNet}
% To enable robust and continuous tracking with DART, we provide hand pose as well as joint angle priors to the optimisation. In the first stage, we used the glove to provide hand pose priors and at the same time used the robust tracking to generate data for naked human hand. 

\paragraph{\textbf{Second Phase}} It is desirable to free the user from having to wear glove in future for any teleoperation. While the first phase operates on RGB image, the second phase operates directly on fused point cloud of bare human hand obtained by back-projecting four depth maps from extrinsically calibrated cameras into a global reference frame. The annotations for this phase come from the data generted in the first phase. Since the camera also provides synchronised depth images, tracking results of the first phase can provide annotations for point clouds. 

The fused point cloud contains both the points on table as well as human body and arm it becomes imperative to first localise the hand. Points on the table are removed by fitting a plane and the remaining points --- containing the arm and human body --- are fed to a PointNet++ based \cite{Qi:etal:arXiv17} architecture that localises the hand as well as provides the hand pose. Our network is based on \cite{Ge:etal:ECCV2018} who estimate hand pose via a voting based regression to the 3D positions of specified keypoints on the hand, a technique reminiscent of spatial-softmax often used in 2D keypoint localisation. It is trained to predict 3D coordinates of 23 keypoints specified on the hand --- 4 joint keypoints each on 5 fingers and 3 at the back of the palm for hand pose estimation. The loss function is standard Euclidean loss between predicted and the ground truth keypoints together with the voting loss inspired by \cite{Ge:etal:ECCV2018}. An auxiliary segmentation loss is also added to obtain hand segmentation. For efficiency reasons, any input point cloud of size $N \times 3$ is sub-sampled uniformly to a fixed $8192 \times 3$ size before feeding to our network. 

\begin{figure*}[t]
    \centering
    \includegraphics[width=1.0\linewidth]{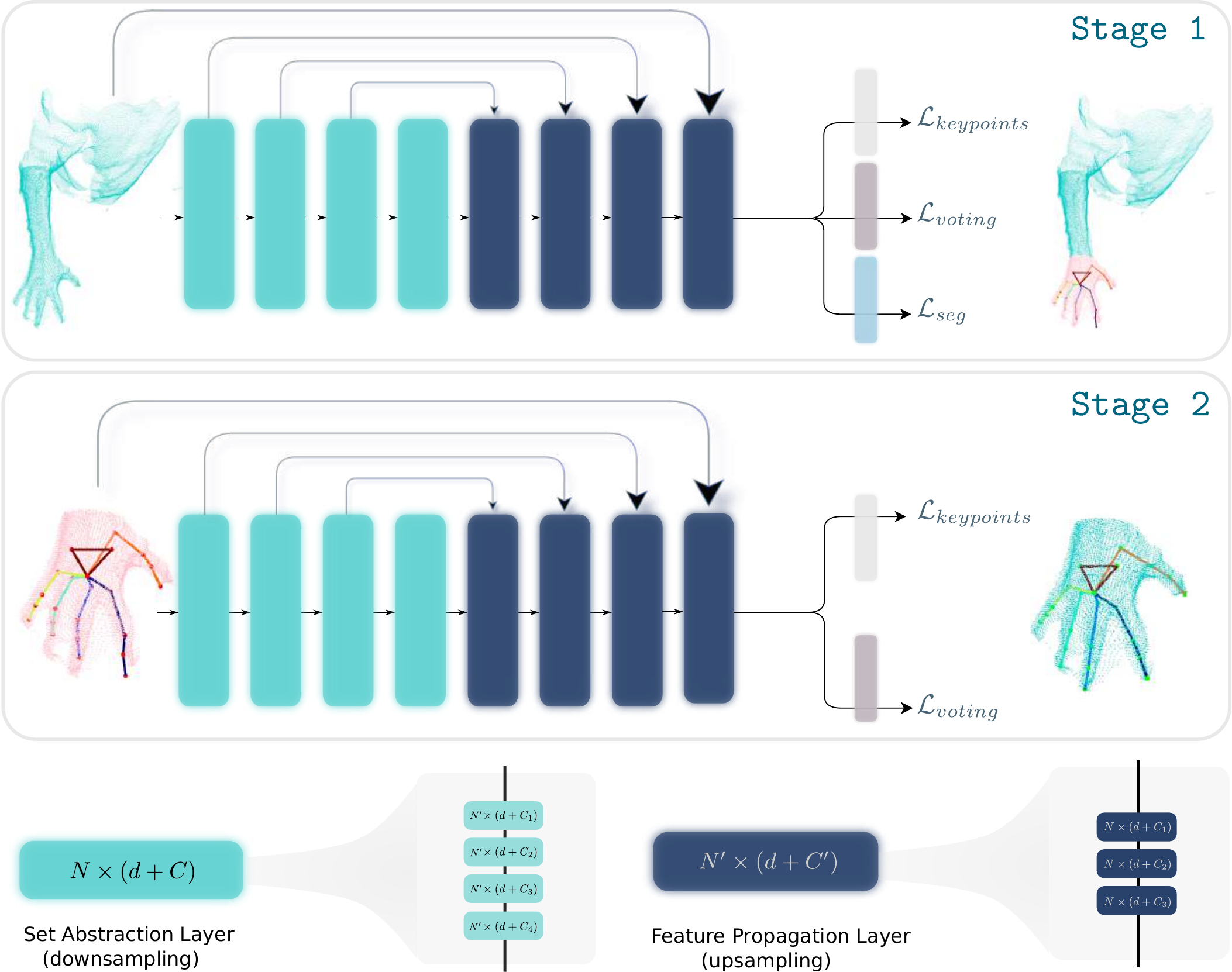}
    %\vspace{-5mm}
    \caption{The PointNet++ inspired architecture operates in two stages. The first stage segments the hand (as shown in pink colour) as well as provides a rough hand pose. The second stage refines the hand pose given the hand segmentation and pose from the first stage. The loss functions include the segmentation loss, the Euclidean loss between the predicted keypoints and ground truth keypoints, and the voting loss as used in \cite{Ge:etal:ECCV2018}. Since the second stage refines keypoints, the segmentation loss is not needed. The set abstraction takes an input of size $N \times (d+C)$ and outputs $N' \times (d+C_4)$ while the feature propagation layer takes $N' \times (d+C')$ input and outputs a tensor of size $N \times (d+C_3)$. Together these two form the backbone of the network. MLPs are used to map the embeddings of PointNet++ backbone to the corresponding desired outputs. More details of the network are in the Appendix.}
    \label{fig:pointnet_architecture}
\end{figure*}

\begin{figure}
  \centering
  \includegraphics[width=\linewidth]{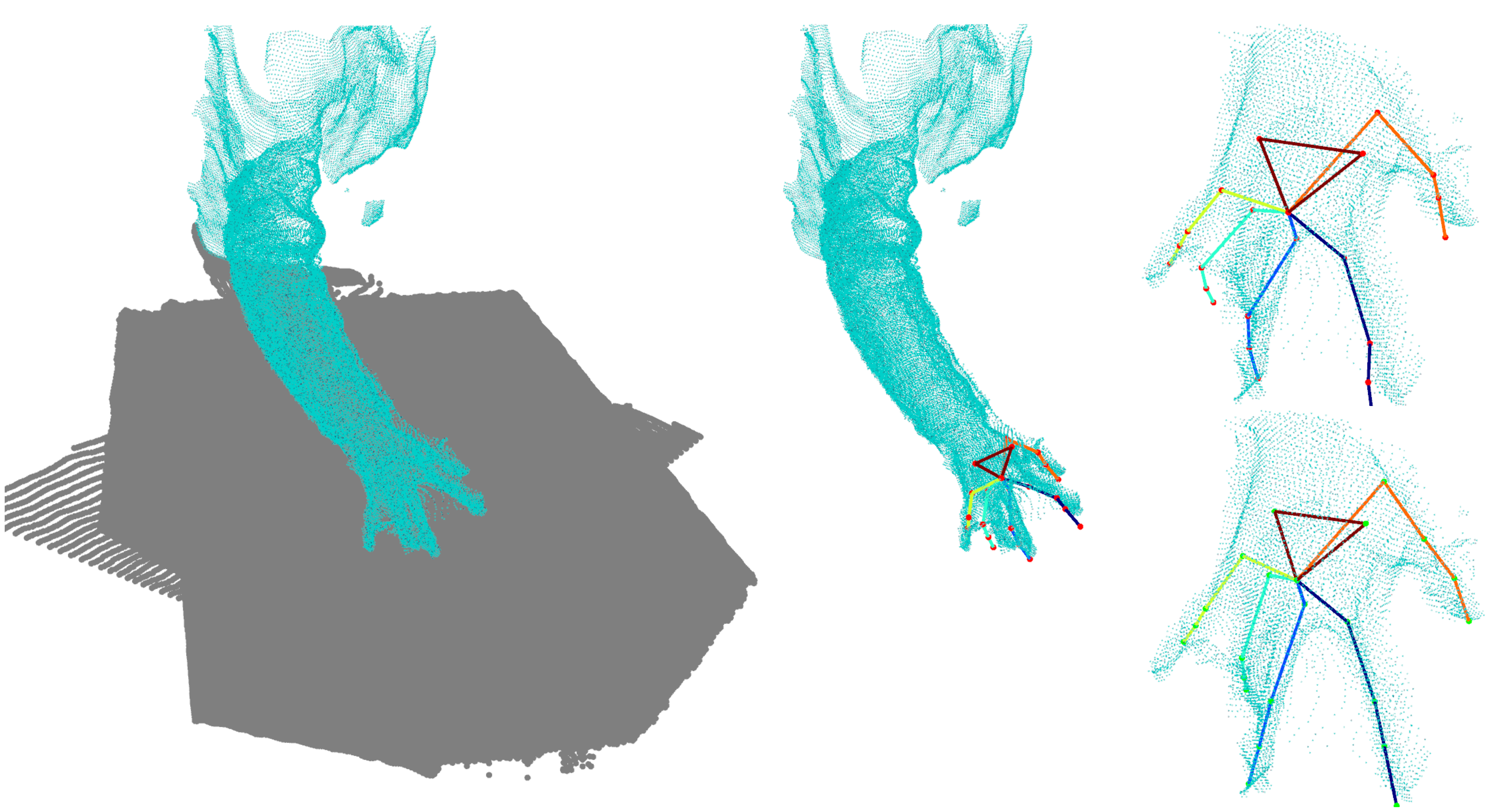}
  \vspace{-5mm}
  \caption{\footnotesize{The input point cloud has points both from the table as well as the human body and arm. A plane was fit to remove points on the table and the remaining points are input to the first stage of our network that recovers the pose of the hand. The second stage refines the pose and provides a more accurate result. The hand images on the right show the result from stage 1 (above) and stage 2 (below).}}
  \label{fig:pointnet}
\end{figure}

While reasonable hand pose estimation and segmentation is achieved, getting high quality predictions for the 20 joint keypoints on the fingers remains difficult with this network. The uniform sub-sampling used at the input means that points on the fingers are not densely sampled and therefore a second stage refinement is needed which resamples points on the hand from the original raw point cloud given the pose and segmentation of the first stage. The overall network architecture is shown in Fig. \ref{fig:pointnet_architecture}. The second stage is trained on only the loss functions pertaining to the keypoints and no segmentation is needed. It uses the points sampled on the hand instead and predicts accurately the 23 keypoints. To enable robustness to any inaccuracies in the hand pose from the first stage, additional randomization is added to the hand pose for second stage. The Fig. \ref{fig:pointnet} shows how the second stage refinement improves the system. Overall, both stages of our network are trained on 100K point clouds collected over a batch of 30-45 minutes each for 7-8 hours in total by running DART with priors from glove. Together they provide annotations for keypoints, joint angles and segmentation. The training takes 15 hours in total on a single NVIDIA TitanXp GPU. 

While keypoints are a natural representation for Euclidean space as used in PointNet++ architectures, most articulated models use joints as a natural parameterisation. Therefore, it is desirable to have output in joint space which can serve as joint priors to DART. A third neural network is trained that maps 23 keypoint locations predicted by our PointNet++ inspired architecture to corresponding joint angles. This neural network, called JointNet, is a two-layer fully connected network that takes input of size $23 \times 3$ and predicts $20$-dimensional vector of joint angles for fingers. 

The neural networks are trained on data collected within the limits of the studio work volume across multiple human hands, ensuring accurate pose fits for this application and enabling sensible priors for DART. Qualitatively, the hand tracker worked well for hands geometrically close to the DART human hand model.
Overall, average keypoint error on a validation set of seven thousand images of differing hand poses and finger configurations was 9.7 mm (comparable to results reported in \cite{Ge:etal:ECCV2018}, but on a different dataset) and joint error was 1.33 degrees per joint.

\section{Robot Motion Generation}
\subsection{Kinematic Retargeting}
\label{sec:kinematic_retargeting}
\begin{figure}
  \vspace{-3mm}
  \centering
  \includegraphics[width=.8\linewidth]{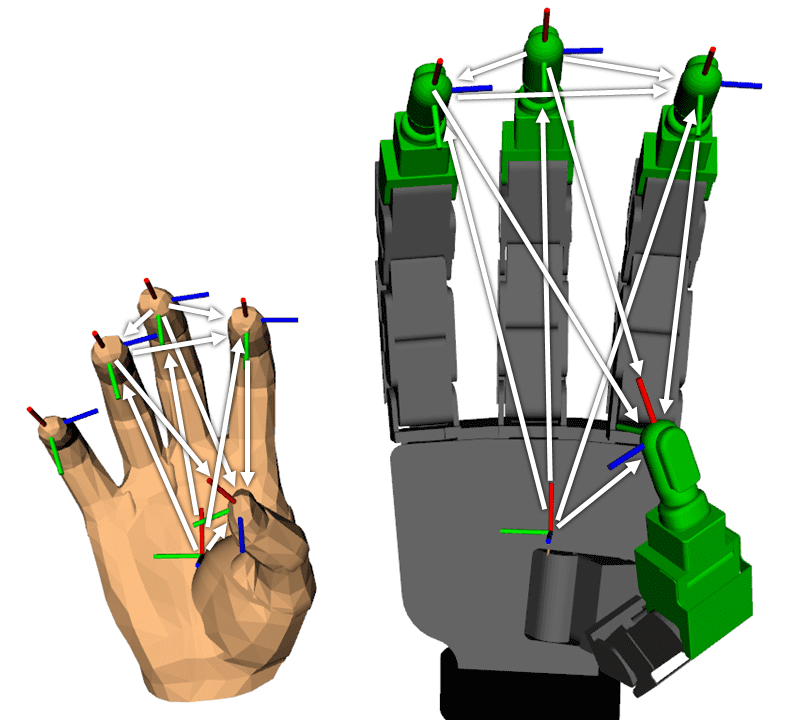}
  \vspace{-2mm}
  \caption{\footnotesize{Task space vectors between fingertips and palm for both the human hand model and Allegro hand used for retargeting optimization.}}
  \label{fig:retargeting_vectors}
\end{figure}

\begin{figure}
  \centering
  \includegraphics[width=\linewidth]{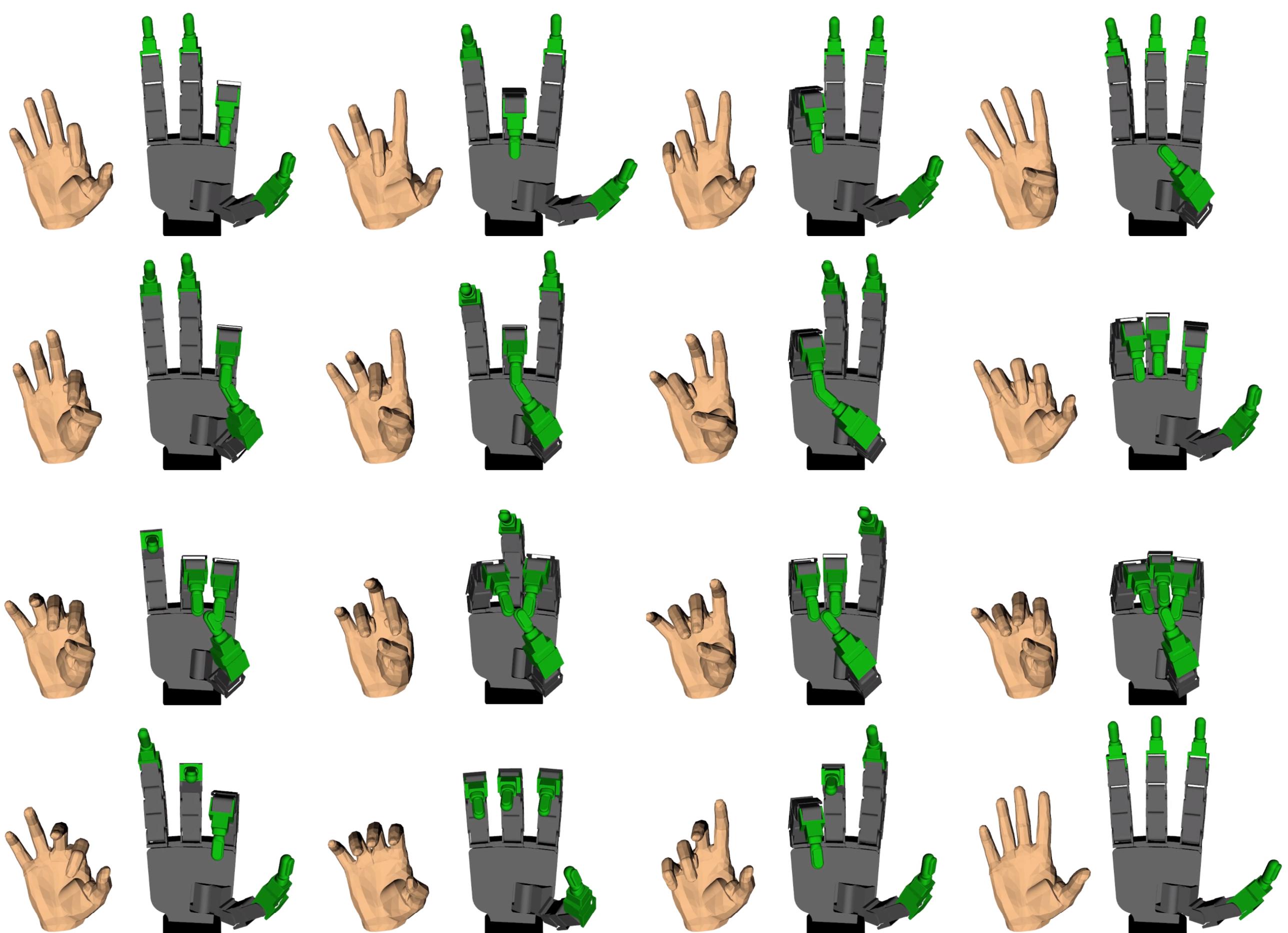}
  \vspace{-3mm}
  \caption{\footnotesize{Canonical kinematic retargeting results between the human hand model and the Allegro hand model.}}
  \label{fig:retargeting_results}
\end{figure}

Teleoperation of a robotic hand that is kinematically disparate from the human hand required a module that can map the observed human hand joints to the Allegro joints. There are many different approaches to kinematic retargeting in the literature.  For instance, a BioIK solver was used to match (between the human and Shadow hand) the positions from palm to the fingertips and medial joints, and the directionality of proximal phalanges and thumb distal phalange in \cite{li:icra2019}. The optimized mapping was used to label human depth images to learn end-to-end a deep network that can ingest a depth image and output joint angles for the Shadow hand. Although interesting, the result produced retargeting results that are not useful for precision grasps (e.g., pinching) where gaps between fingertips need to be small or zero. Motion retargeting is also present in the animation field. For instance, a deep recurrent neural network was unsupervised trained to retarget motion between skeletons \cite{villegas2018neural}. Although the synthesized motion look compelling, it is unclear whether these approaches work well enough for teleoperative manipulation where the important task spaces for capturing grasping and manipulation behavior are likely not equivalent to those that capture visual consistencies. The approach herein prioritized fingertip task-space metrics because distal regions are of highest priority in grasping and manipulation tasks as measured by their contact prevalence \cite{sundaram2019learning}, degree of innervation \cite{johansson2009coding}, and heightened controllability for fine, in-hand manipulation skill \cite{van2018comparative}. Moreover, the joint axes and locations between the two hands are strikingly different, and therefore, no metrics directly comparing joint angles between the two hands are used. To capture and optimize for the positioning of fingertips, both distance and direction among fingertips were considered. Specifically, the cost function for kinematic retargeting was chosen as

\begin{equation*}
\mathcal{C}(q_h, q_a) = \frac{1}{2} \sum_{i=0}^{\mathsf{N}} s(d_i) ||r_i (q_a) - f(d_i) \hat{r}_i(q_h)|| ^ 2 + \gamma ||q_a||^2,
\end{equation*}

\noindent where $q_h, q_a$ are the angles of the human hand model and Allegro hand, respectively, $r_i \in \mathbb{R}^3$ is the vector pointing from the origin of one coordinate system to another, expressed in the origin coordinate system (see Fig. \ref{fig:retargeting_vectors}). Furthermore, $d_i = ||r_i(q_h)||$ and $\hat{r}_i(q_h) = \frac{r_i(q_h)}{||r_i(q_h)||}$. The switching weight function $s(d_i)$ is defined as

\begin{equation*}
s(d_i) = 
\begin{cases}
    1, & d_i > \epsilon \\
    200, & d_i \leq \epsilon \hspace{2mm} \land \hspace{2mm} r_i(q_h) \in \mathcal{S}_1 \\
    400, & d_i \leq \epsilon \hspace{2mm} \land \hspace{2mm} r_i(q_h) \in \mathcal{S}_2, \\
\end{cases}
\end{equation*}

\noindent where $\mathcal{S}_1$ and $\mathcal{S}_2$ are vector sets defined in Table \ref{tab:vector_sets}, The distancing function, $f(d_i) \in \mathbb{R}$, is defined as

\begin{equation*}
f(d_i) = 
\begin{cases} 
      \beta \, d_i, & d_i > \epsilon \\
      \eta_1, & d_i \leq \epsilon \hspace{2mm} \land \hspace{2mm} r_i(q_h) \in \mathcal{S}_1 \\
      \eta_2, & d_i \leq \epsilon \hspace{2mm} \land \hspace{2mm} r_i(q_h) \in \mathcal{S}_2, \\
\end{cases}
\end{equation*}

\noindent where $\beta = 1.6$ is a scaling factor, $\eta_1 = 1 \times 10^{-4} \, m$ closes the distance between a primary finger and the thumb, and $\eta_2 = 3 \times 10^{-2} \, m$ forces a minimum separation distance between two primary fingers when both primary fingers are being projected with the thumb. These projections ensure that contact between primary fingers and the thumb are close without inducing primary finger collisions in a precision grasp. This was found to be particularly useful in the presence of visual finger tracking inaccuracies.  Importantly, the vectors $r_i$ not only capture distance and direction from one task space to another, but their expression in local coordinates further contains information on how the coordinate systems, and thereby fingertips, are oriented with one another. The coordinate systems of the human hand model must therefore have equivalent coordinate systems on the Allegro model with similarity in orientation and placement. The vectors shown in Fig. \ref{fig:retargeting_vectors} were a minimal set that produced the desired retargeting behavior. Finally, $\gamma = 2.5 \times 10^{-3}$ is a weight on regularizing the Allegro angles to zero (equivalent to fully opened the hand). This term helped greatly with reducing redundancy in solution and ensured that the hand never entered strange minima that was difficult to recover from (e.g., the fingers embedding themselves into the palm). Finally, to further reduce redundancy, the search space, and to emulate the human hand, the distal joints for the primary (index, middle, and ring) fingers of the Allegro hand were constrained to equal their medial joints. Various mappings from human hand to Allegro as produced by our kinematic retargeting are show in Fig. \ref{fig:retargeting_results}.

For implementation, the above cost function was minimized in real-time using the Sequential Least-Squares Quadratic Programming (SLSQP) algorithm \cite{Kraft:1988, Kraft:1994} of the NLopt library \cite{NLOpt}. The routine was initiated with Allegro joint angles set to zero, and every solution thereafter was initiated with the preceding solution. Moreover, the forward kinematic calculations between the various coordinate systems of both the human hand model and Allegro hand were found using the Orocos Kinematics and Dynamics library \cite{OrocosKDL}. Finally, a first-order low-pass filter was applied to the raw retargeted joint angles in order to remove high-frequency noise present in tracking the human hand and to smooth discrete events like the projection algorithm inducing step-response changes in retargeted angles.

\begin{table}[]
\caption{Description of vector sets used in kinematic retargeting.}
\centering
\begin{tabular}{|p{1cm}|p{6.7cm}|}
\hline
\textbf{Set}  & \textbf{Description} \\ \hline
$\mathcal{S}_1$ & Vectors that originate from a primary finger (index, middle, ring) and point to the thumb. \\ \hline
$\mathcal{S}_2$ & Vectors between two primary fingers when both primary fingers have associated vectors $\in \, \mathcal{S}_1$, \textit{e.g.}, both primary fingers are being projected with the thumb. \\ \hline
\end{tabular}
\label{tab:vector_sets}
\end{table}

\subsection{Riemannian Motion Policies}
\label{sec:RMPs}
Riemannian Motion Policies (RMPs) are real-time motion generation methods that calculate acceleration fields from potential function gradients and corresponding Riemannian metrics \cite{ratliff2018riemannian, cheng2018rmpflow}. RMPs combines the generation of multi-priority Cartesian trajectories and collision avoidance behaviors together in one cohesive framework. They are used to control the Cartesian pose of the Allegro palm given the observed human hand pose while avoiding arm-palm collisions with the table or operator using collision planes. Given these objectives, the RMPs generated target arm joint trajectories that were sent to the arm's torque-level impedance controller at 200 Hz. The kinematically retargeted Allegro angles were sent to the torque-level joint controller at 30 Hz. One final calibration detail involves registering human hand pose movements with the robot system. This was accomplished by finding the transformation from the robot coordinate system to the camera coordinate system. This transformation was calculated using the initial view of the human hand and an assumed initial pose of the robot hand. To facilitate spatial reasoning of the pilot, the desired initial hand pose of the pilot is a fully open hand with the palm parallel to the table and fingers pointing forwards. The assumed initial pose of the robot mimics this pose. In this way, the robot moves in the same direction as the pilot's hand, enabling intuitive spatial reasoning.

\section{Experiments}
\begin{figure*}
  \centering
  \includegraphics[width=\linewidth]{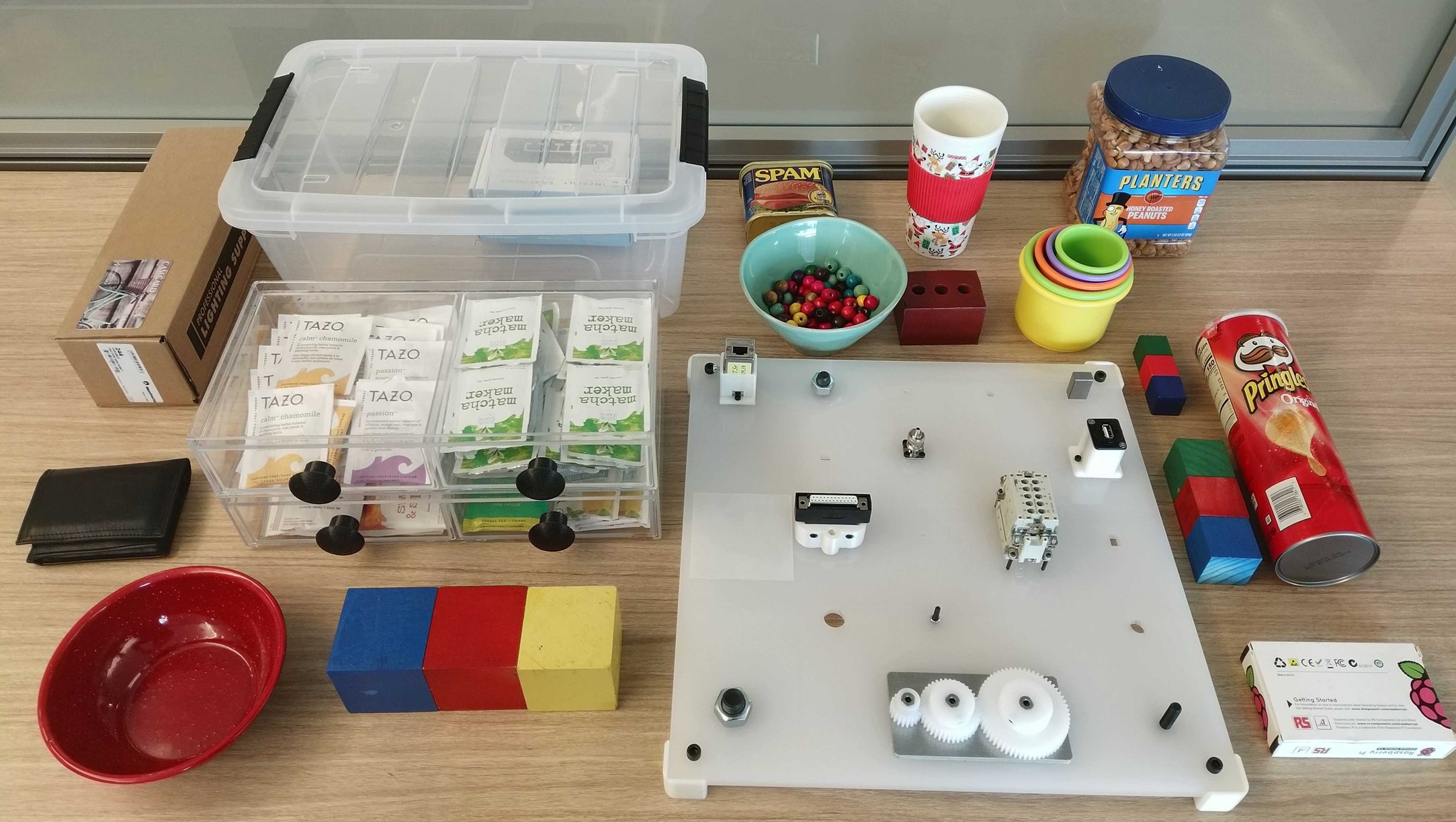}
  \vspace{-5mm}
  \caption{\footnotesize{Objects used for teleoperation. Various tasks performed with these objects can be viewed at \url{https://sites.google.com/view/dex-pilot}.}}
  \label{fig:test_objects}
\end{figure*}

\begin{figure*}
  \centering
  \includegraphics[width=\linewidth]{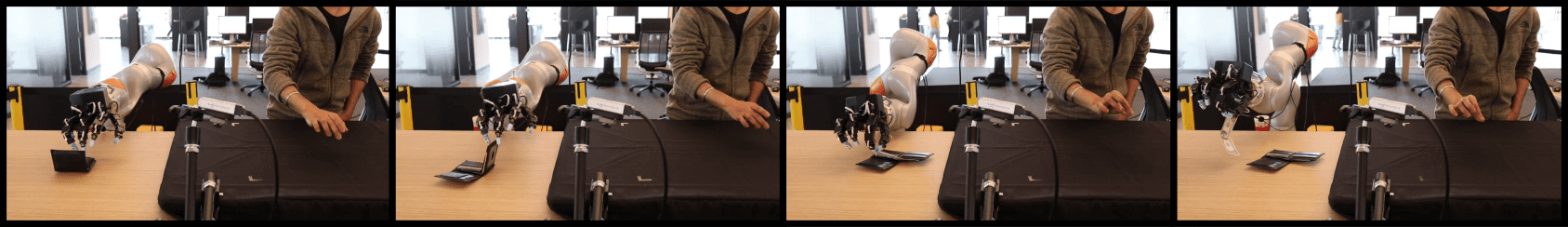}
  \vspace{-5mm}
  \caption{Extracting money from a wallet. The pilot has to open the wallet first and move it to a particular vantage location in order to pull out paper currency. Importantly, the hand is able to keep the paper by pinching fingers against each other.}
  \label{fig:extracting_money}
\end{figure*}

\begin{figure*}
  \centering
  \includegraphics[width=\linewidth]{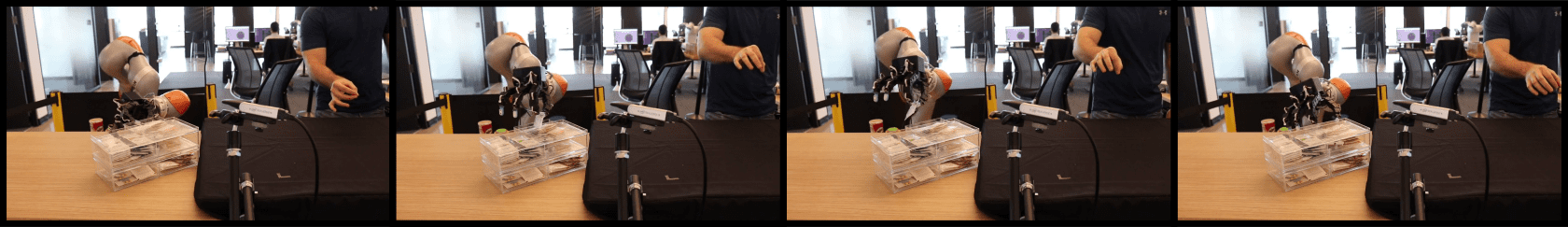}
  \vspace{-5mm}
  \caption{Opening a tea drawer, extracting a tea bag, and closing the drawer. This is a somewhat long horizon task and requires dexterity in opening the drawer and holding on to the tea bag.}
  \label{fig:extracting_tea}
\end{figure*}

\begin{figure*}
  \centering
  \includegraphics[width=\linewidth]{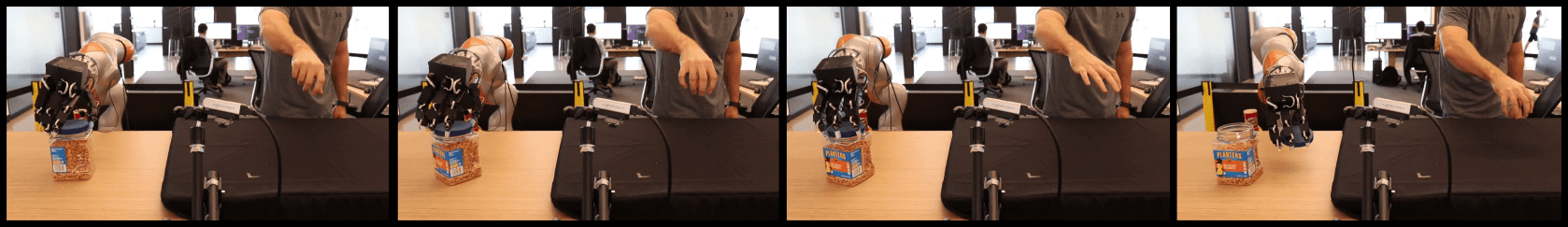}
  \vspace{-5mm}
  \caption{Opening a peanut jar. The task needs rotating the cap multiple times in order to open while maintaing the contacts.}
  \label{fig:opening_jar}
\end{figure*}

The DexPilot system was systematically tested across a wide range of physical tasks that test precision and power grasps, prehensile and non-prehensile manipulation, and finger gaiting (see Fig. \ref{fig:test_objects} for test objects and Figs. \ref{fig:extracting_money}, \ref{fig:extracting_tea}, \ref{fig:opening_jar} for teleoperative manipulation). The description of the tasks are provided in Table \ref{tab:dataset}.

\begin{table*}[t]
\caption{The test suite consists of 15 different tasks of varying complexity ranging from classic pick and place to multi-step, long horizon tasks. Each of these tasks is operated with 5 consecutive trials to avoid preferential selection and success rate is reported accordingly. If the object falls out of the workspace volume the trial is considered a failure. The last column represents the skills needed for teleoperation as the hand changes its state over time.}
    \centering
        \begin{tabular}{>{\raggedright}p{4.0cm} | p{6.6cm} | p{6.0cm}}
        \hline
        \textbf{Task} & \textbf{Description} & \textbf{Required Skills}\\
        \hline
        \rowcolor{LightCyan}
        \texttt{Pick and Place} 
        \begin{itemize}
            \item \texttt{Foam brick}
            \item \texttt{Pringles can} 
            \item \texttt{Spam box}
        \end{itemize} & Pick object on the table and place it in a red bowl. & grasping, releasing   \\
        % \texttt{pick and place (pringles)} & Pick Pringles box and place in red bowl standing straight & 5m   \\
        % \rowcolor{LightCyan}
        % \texttt{pick and place (spam box)} & Pick spam can and place in red bowl & 5m   \\ 
        \texttt{Block Stacking} \begin{itemize}
            \item \texttt{Large (L)  (6.3cm)}
            \item \texttt{Medium(M) (3.8cm)} 
            \item \texttt{Small (S) (2.3cm)}
        \end{itemize}& Stacking three blocks on top of each other. & precision grasping, precision releasing   \\ 
        \rowcolor{LightCyan}
        \texttt{Pouring Beads} & Pour beads from a cup into a bowl. & grasping, pouring   \\ 
        % \rowcolor{LightCyan}
        \texttt{Opening Jar} & Open peanut jar and place lid on table. & finger gaiting, grasping, releasing   \\
        \rowcolor{LightCyan}
        \texttt{Brick Gaiting} & Pick up and in-hand rotate brick 180 degrees and place back down. & grasping, in-hand manipulation, releasing   \\ 
        % \rowcolor{LightCyan}
        \texttt{Container} & Open plastic container, extract and open cardboard box. & twisting, pulling, pushing, grasping, in-hand manipulation   \\ 
        % \texttt{block\_stacking} & Stacking three blocks on top of each other & 5m   \\ 
        \rowcolor{LightCyan}
        \texttt{Cup Insertion} & Inserting concentric cups inside each other. & grasping, releasing   \\ 
        \texttt{Tea Drawer} &  Pull open tea drawer, extra single bag of tea and place on table, close tea drawer. & precision grasping, pulling, pushing, releasing\\
        \rowcolor{LightCyan}
         \texttt{Card Sliding} & Slide a card along the box and pick it up with two fingers.  & sliding, precision grasping, releasing\\
         \texttt{Wallet} & Open the wallet and pull out paper money. & precision grasping, pulling, pushing, in-hand manipulation\\
         \rowcolor{LightCyan}
        \texttt{Box Flip} & Flip the box by 90 degrees and place it on the designated goal. & pushing, grasping, releasing\\
        \hline
        \end{tabular}
\label{tab:dataset}
\end{table*}

\begin{figure}
  \centering
  \includegraphics[width=.9\linewidth]{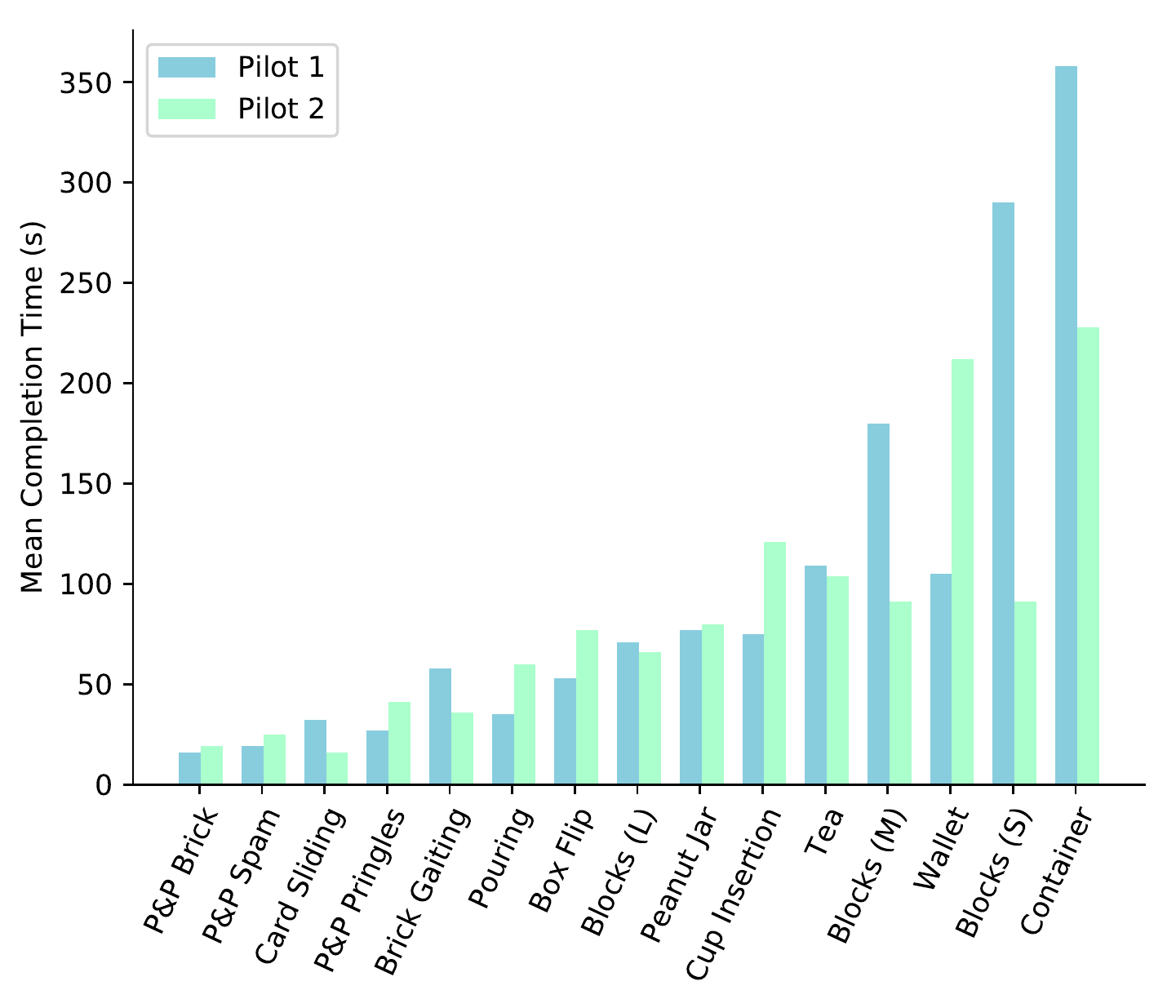}
  \vspace{-4mm}
  \caption{Mean completion time of teleoperation tasks across two pilots run over five successive trials without reset.}
  \label{fig:ct_bar_plots}
\end{figure}

\begin{figure*}
    \centering
    \includegraphics[width=1.0\linewidth]{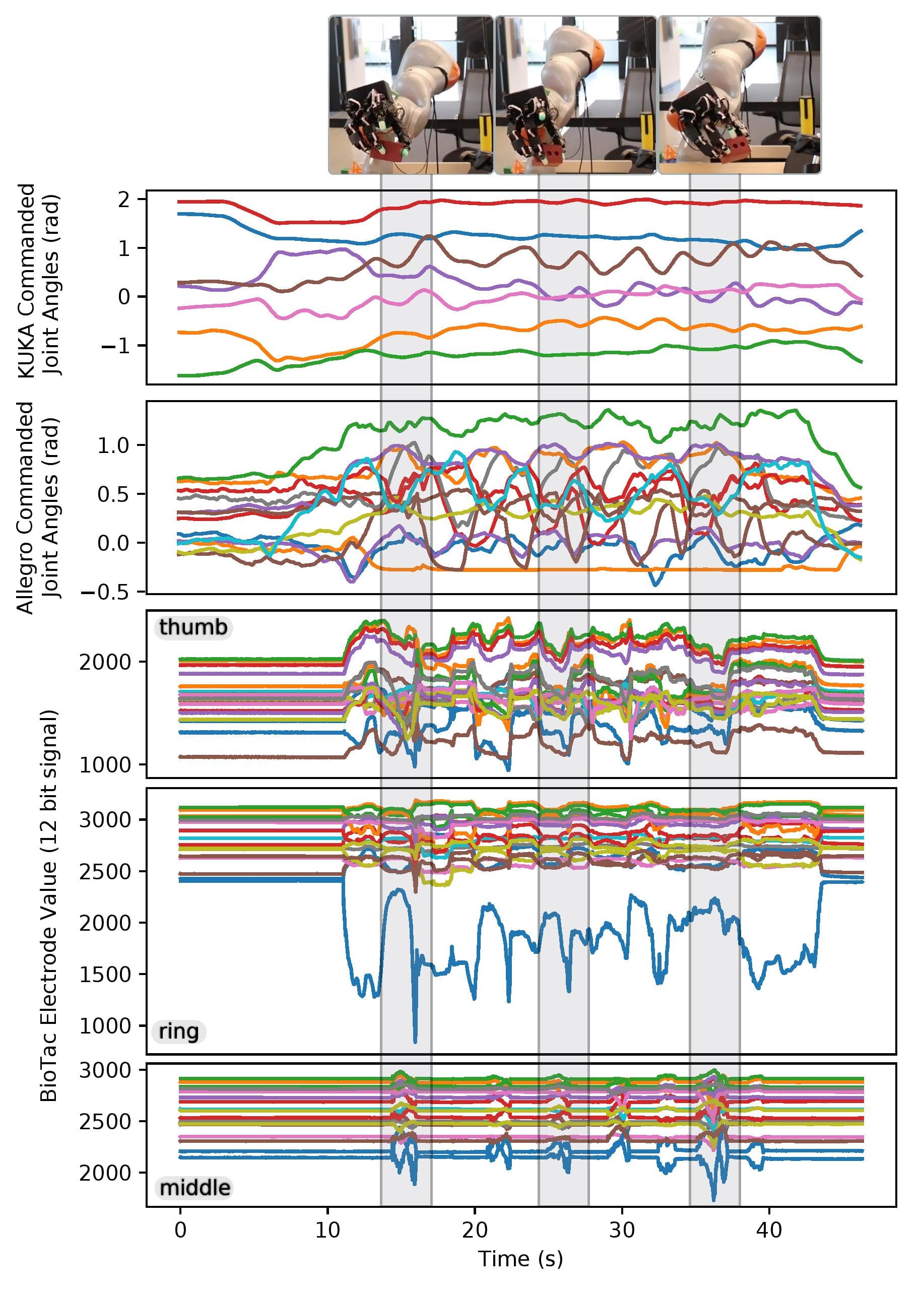}
    \vspace{-10mm}
    \caption{BioTac tactile signals and robot joint position commands during the brick gaiting task where the middle finger makes the contact with the brick a total of 7 times over a 40 second duration in order to rotate it by 180 degrees. The 7 contacts made are also evident in the BioTac signals of the middle finger. The thumb and the ring finger remain pinched in order to hold the brick in hand. }
    \label{fig:biotac_img_sync}
\end{figure*}

Before benchmarking, the pilots went through a warm-up training phase where they tried to solve the task with 3-5 non-consecutive attempts. Later, five consecutive test trials were conducted by each pilot for each task to avoid preferential selection of results and pilots were graded based on their performance. The performance metrics for these tasks include mean completion time (CT) and success rate which capture speed and reliability of the teleoperation system. The system was tested with two pilots and the performance measures are reported in Fig. \ref{fig:ct_bar_plots} (see more videos at \url{https://sites.google.com/view/dex-pilot}). Overall, the system can be reliably used to solve a variety of tasks with a range of difficulty. Differences in mean CT across tasks indicate the effects of task complexity and horizon scale. Discrepancies in mean CT across pilots per task indicate that there does exist a dependency on pilot behavior. Effects include fatigue, training, creativity, and motivation. The ability to solve these tasks reveal that the teleoperation system has the dexterity to exhibit precision and power grasps, multi-fingered prehensile and non-prehensile manipulation, in-hand finger gaiting, and compound in-hand manipulation (\textit{e.g.}, grasping with two fingers while simultaneously manipulating with the remaining fingers). Note, certain tasks, \textit{e.g.} \texttt{Container} and \texttt{Wallet}, take a particularly long time to teleoperate largely due to the fact that these tasks are multi-stage tasks. On the other hand, the task requiring picking small cubes is particularly challenging because the behavior of releasing the grasps on these objects with the projection scheme used in kinematic retargeting can be unpredictable. Nevertheless, such a rich exhibition of dexterous skill transferred solely through the observation of the bare human hand provides empirical evidence that the approach and architecture herein works well. An important aspect that is worth highlighting is that although the full teleoperation process for a particular task may not be perfect (\textit{e.g.} the pilot may lose an object in hand but fetches it again to accomplish the task), the data collected is still equally valuable in helping robot learn to recover from failures. Additionally, in the spirit of \cite{Lynch:etal:CoRL2019}, the data can be regarded as play data which is useful to learn long range planning. Visualization of a sensorimotor solution to the \texttt{Brick Gaiting} task can be seen in Fig. \ref{fig:biotac_img_sync}. As shown, discrete events like intermittent finger-object contacts can be observed in the tactile signals. Undulations in these state-action signals reveal the rich, complex behavior evoked in the system through this embodied setting. Force estimates can also be obtained as in \cite{Sundaralingam:etal:ICRA2019}. This data can now be generated on-demand for a particular task with the hope that functional sensorimotor patterns may be gleaned and imparted to the system in an autonomous setting.

\begin{figure}
  \centering
  \includegraphics[width=.9\linewidth]{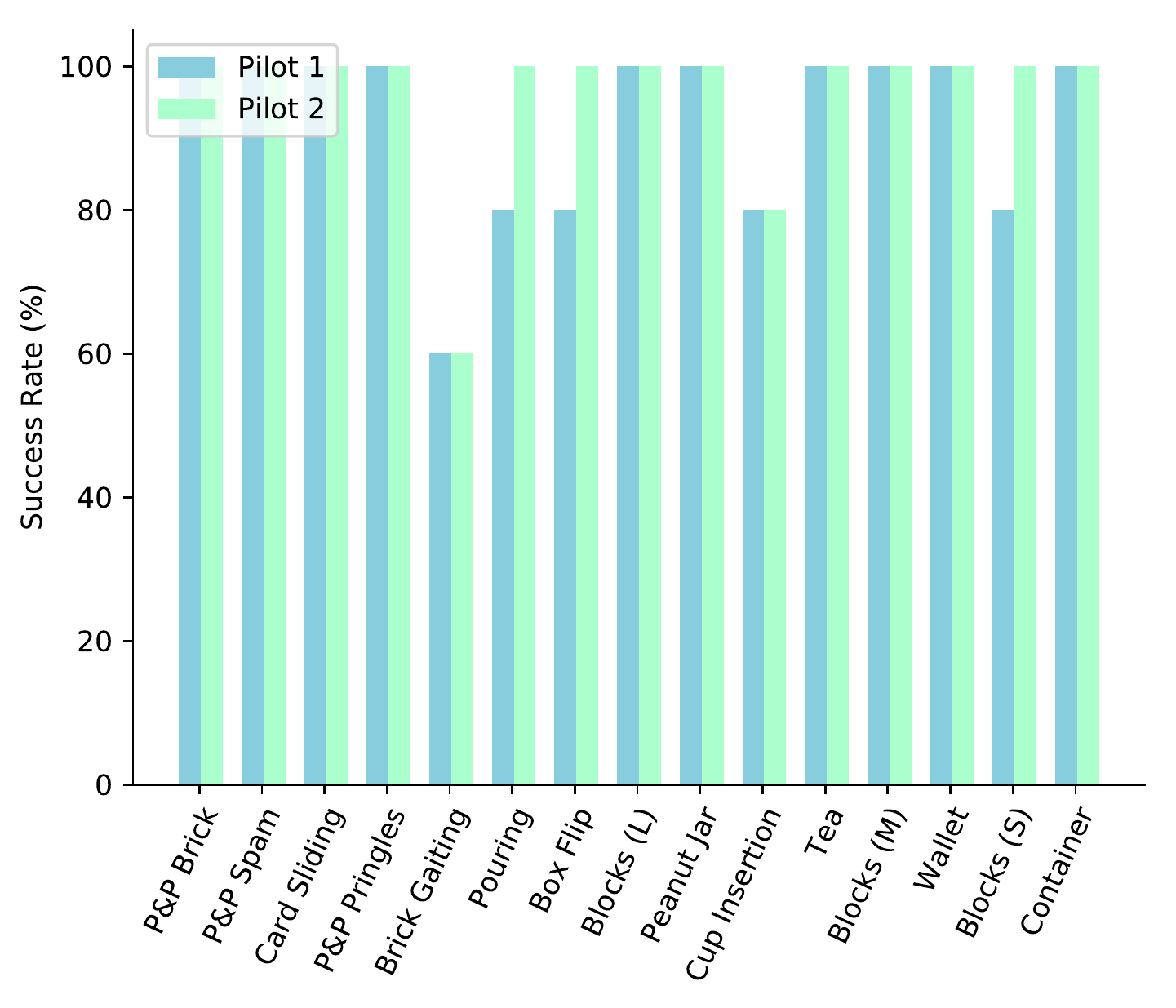}
  \vspace{-4mm}
  \caption{Success rate of teleoperation tasks across two pilots run over five successive trials without reset.}
  \label{fig:sr_bar_plots}
\end{figure}

\begin{figure*}[htp]
  \centering
  \includegraphics[width=\linewidth]{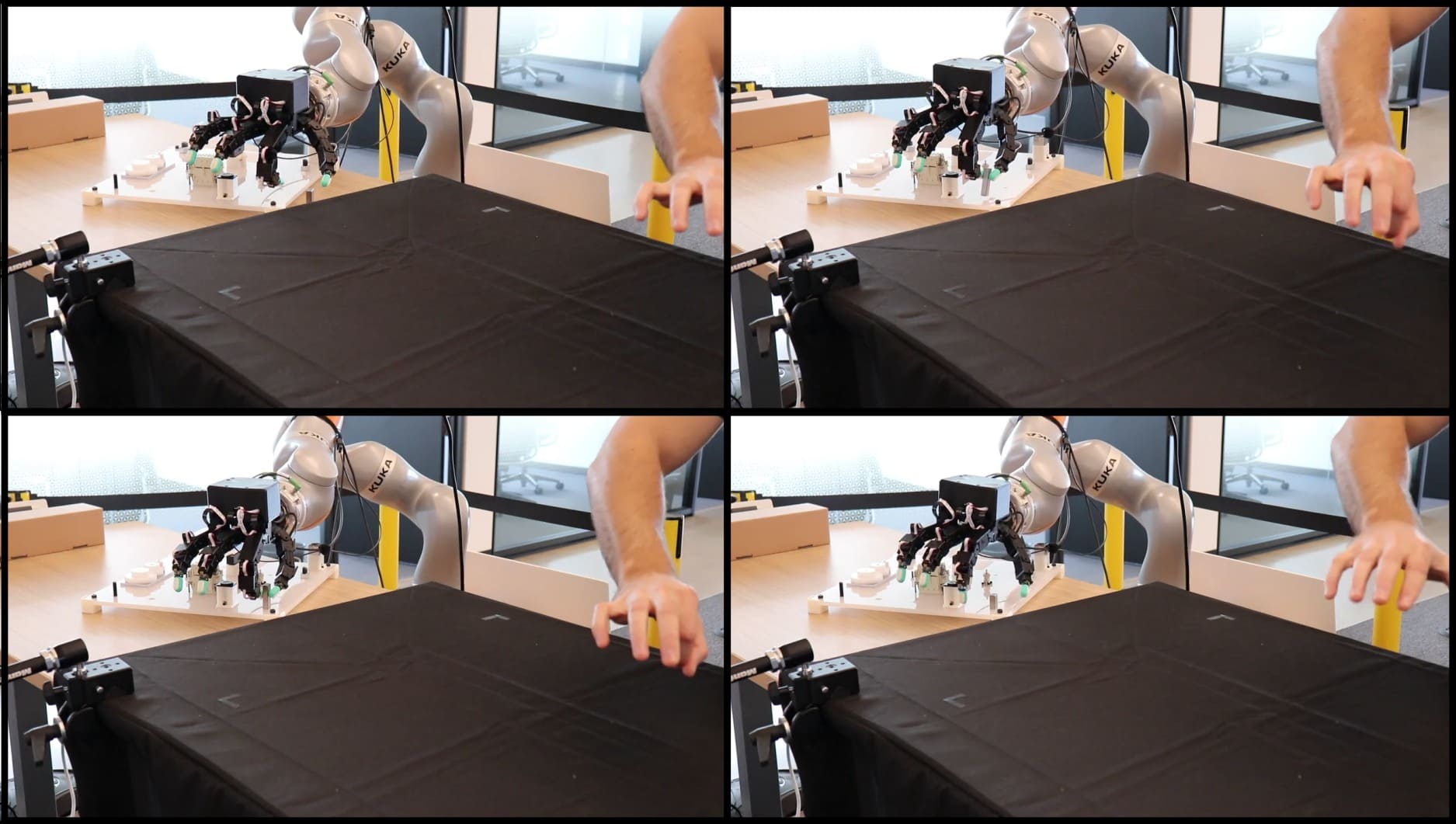}
  \vspace{-4mm}
  \caption{NIST task board peg-in-hole insertion. The peg dimensions are 16 $mm$ $\times$ 10 $mm$ $\times$ 49.5 $mm$ with the hole clearance of 0.1 $mm$. } 
  \label{fig:nist_board}
\end{figure*}

\begin{figure*}
  \centering
  \includegraphics[width=\linewidth]{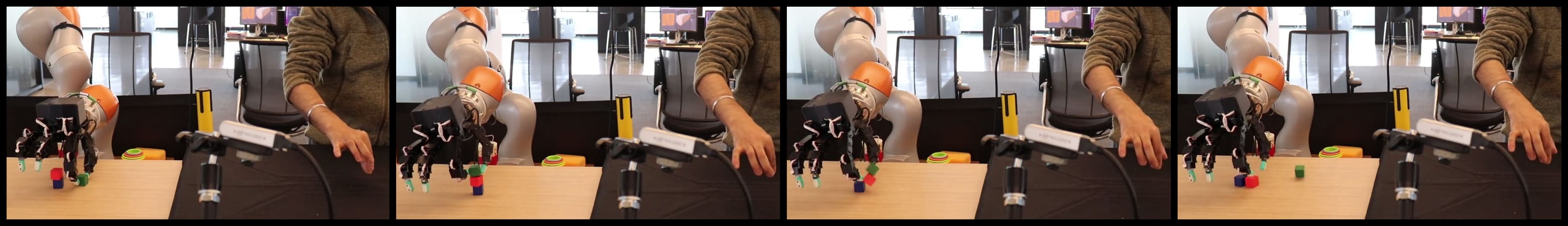}
  \vspace{-5mm}
  \caption{The projection scheme used in the kinematic retargeting that compensates for inaccuracies in the hand tracking can make releasing small objects from the fingers difficult leading to losing objects often.}
  \label{fig:failure_case}
\end{figure*}

\section{Discussion}
DexPilot enabled a highly-actuated hand-arm system to find a motor solution to a variety of manipulation tasks by translating observed human hand and finger motion to robot arm and finger motion. Importantly, several tasks like extracting money from a wallet and opening a cardboard box within a plastic container were so complex that hand-engineering a robot solution or applying learning methods directly are likely intractable. Solving these tasks and the others through the embodied robotic system revealed that a sensorimotor solution did exist and that these solutions can be generated on-demand for many demonstrations. Furthermore, creating these solutions on the system itself allows for the reading, access, and storage of the 92 tactile signals in the robot's fingertips, 23 commanded and measured joint position and velocity signals through the hand and arm, 23 torque commands throughout the system, and any camera feeds associated with the system. This rich source of data will be critical for the application of learning methods that may hopefully learn to solve complex, multi-stage, long horizon tasks. Applying these learning methods is a future direction for this enabling work. Moreover, the current DexPilot system will be improved in a variety of ways in the future as well. For instance, human hand tracking accuracy could be improved with advancements in deep learning architectures, inclusion of RGB data, larger data sets, and changes to imaging hardware. Ideally, human hand tracking accuracy should be improved enough to greatly reduce the projection distance in the kinematic retargeting approach, enhancing fine finger control and manipulation over small objects and multi-fingered precision grasps. Grasp and manipulation control algorithms \cite{van2018comparative} could be implemented on the hand that automates force modulating control to reduce the control burden on the user and minimize unintentional part drops from the application of incorrect grip forces. Finally, intent recognition schemes could be implemented that enables the robot to predict human intention and deploy automated solutions, \textit{e.g.}, the system recognizes the human's intent to grasp an object and the system automatically acquires the grasp. Such a co-dependent system would allow a human to direct the robot with full knowledge of a task and its solution strategy, while the robot system controls the finer details of the solution implementation.

\section{Limitations}

Overall, DexPilot is a viable, low-cost solution for teleoperating a high DoA robotic system; however, there are areas that could be improved with the current implementation. For instance, the observable work volume of the pilot could be enlarged to allow for tasks that cover greater distances with better RGB-D cameras. The projection schemes in kinematic retargeting enabled successful manipulation of small objects, but can interfere with finger gaiting tasks and timely releasing grasps on small objects as shown in Fig. \ref{fig:failure_case}. Fortunately, this feature can be turned off when desired, but ultimately, this interference should be non-existent. This issue could be solved entirely with hand tracking that can accurately resolve situations where the human hand fingertips are making contact. Human hand tracking could also be further improved with enhanced robustness across size and shape of the pilot's hand. The lack of tactile feedback makes precision tasks difficult to complete. To compensate, building in autonomous control features could alleviate some of the control burden on the pilot. Furthermore, the system latency could be reduced and the responsiveness of the RMP motion generator could be tuned for faster reactions. Finally, high-precision tasks like slip-fit peg-in-hole insertions pose a challenge to the current system. Peg-in-hole insertions on the NIST task board \cite{van2016robotic, NISTassembly} were attempted with DexPilot, but results were mostly unsatisfactory. Fig. \ref{fig:nist_board} shows one attempt where the pilot managed to successfully guide the system to  insert a 16 mm $\times$ 10 mm $\times$ 49.5 mm peg with a 0.1 mm hole clearance. Encouragingly, these tasks which have not been successfully solved by robotic systems that use machine vision and robotic hands are now made possible by such systems. However, success rates are typically within 10 \% when under specific conditions on the initial placement of the NIST task board and parts (close to the user). The difficulty of completing such tasks could be significantly reduced with improved hand tracking performance, automated precision grip control on the assembly objects, and improved sight to the small parts and insertion locations.

\section{Acknowledgements}
We would like to thank Adithya Murali, Jonathan Tremblay, Balakumar Sundaralingam, Clemens Eppner, Chris Paxton, Yashraj Narang, Alexander Lambert, Timothy Lee, Michelle Lee, Adam Fishman, Yunzhu Li, Ajay Mandlekar, Muhammad Asif Rana, Carlos Florensa, Krishna Murthy Jatavallabhula, Jonathan Tompson, Joseph Xu, Kendall Lowrey, Lerrel Pinto, Ian Abraham, Jan Czarnowski, Raluca Scona, Tucker Hermans, Svetoslav Kolev, Umar Iqbal, Pavlo Molchanov, Emo Todorov, Artem Molchanov, Yevgen Chebotar, Viktor Makoviychuk, Visak Chadalavada and James Davidson for useful discussions and feedback during the development of this work.

% \section{Demonstrations}
% \begin{itemize}
%     \item HaptX \url{https://twitter.com/RobotAndAIWorld/status/1126182331310538753} and \url{https://twitter.com/HaptX/status/1144352016690339840}
%     \item SciRobotics \url{https://twitter.com/SciRobotics/status/1155115838372044801}
%     \item Open Bionics Hand \url{https://www.youtube.com/watch?v=76_xuUS-ba0}
% \end{itemize}

% \section{Comparisons}

% \begin{itemize}
%     \item We need some comparisons to the OpenPose work \url{https://github.com/CMU-Perceptual-Computing-Lab/openpose}. So, I tried it today and is quite fragile. Just need to add a video to show that it doesn't work that well.
%     \item We need comparisons to depth based hand-pose estimation \url{https://github.com/xkunwu/depth-hand}.
%     \item Try deepMimic for various primitive motions \url{https://github.com/xbpeng/DeepMimic}.
%     \item We obtain the hand model from \url{http://mano.is.tue.mpg.de/} and turn it into an OBJ mesh via \url{https://github.com/hassony2/manopth}.
%     \item Kinematic retargetting. 
%     \item Explain RMPs.
% \end{itemize}

\renewcommand*{\bibfont}{\small}
\setlength\bibsep{0.18\baselineskip}
\bibliographystyle{IEEEtran} 
\bibliography{teleop}
\clearpage

\onecolumn
\section{Appendix}

\subsection{GloveNet: Hand Tracking with Colour Glove}

Hand tracking with glove is done via keypoint detection with neural networks. The user wears a black glove with coloured blobs and moves the hand on a table covered with black cloth \textit{i.e.} the scene is instrumented in a way that aids hand tracking. Since the colours are distinct and that most of the background is black, we can use OpenCV HSV colour thresholding to generate annotations for these coloured blobs. The HSV thresholds vary with the time of the day and therefore we collect data across days to build a big dataset of 50K images. We use a neural network to fit this data which makes the whole process robust to lighting changes and bad annotations and avoids the burden on the user to find the appropriate thresholds at test time. The network, called GloveNet, uses 4 layers of ResNet-50 \cite{He:etal:CVPR16} with spatial-softmax at the end to regress to the 2D locations of finger-tips. We choose the recently proposed anti-aliased ResNet-50 from \cite{Zhang:etal:ICML19} for accurate and consistent predictions. We explain various stages of the pipeline below.

\begin{figure*}[htp]
  \centering
  \includegraphics[width=\linewidth]{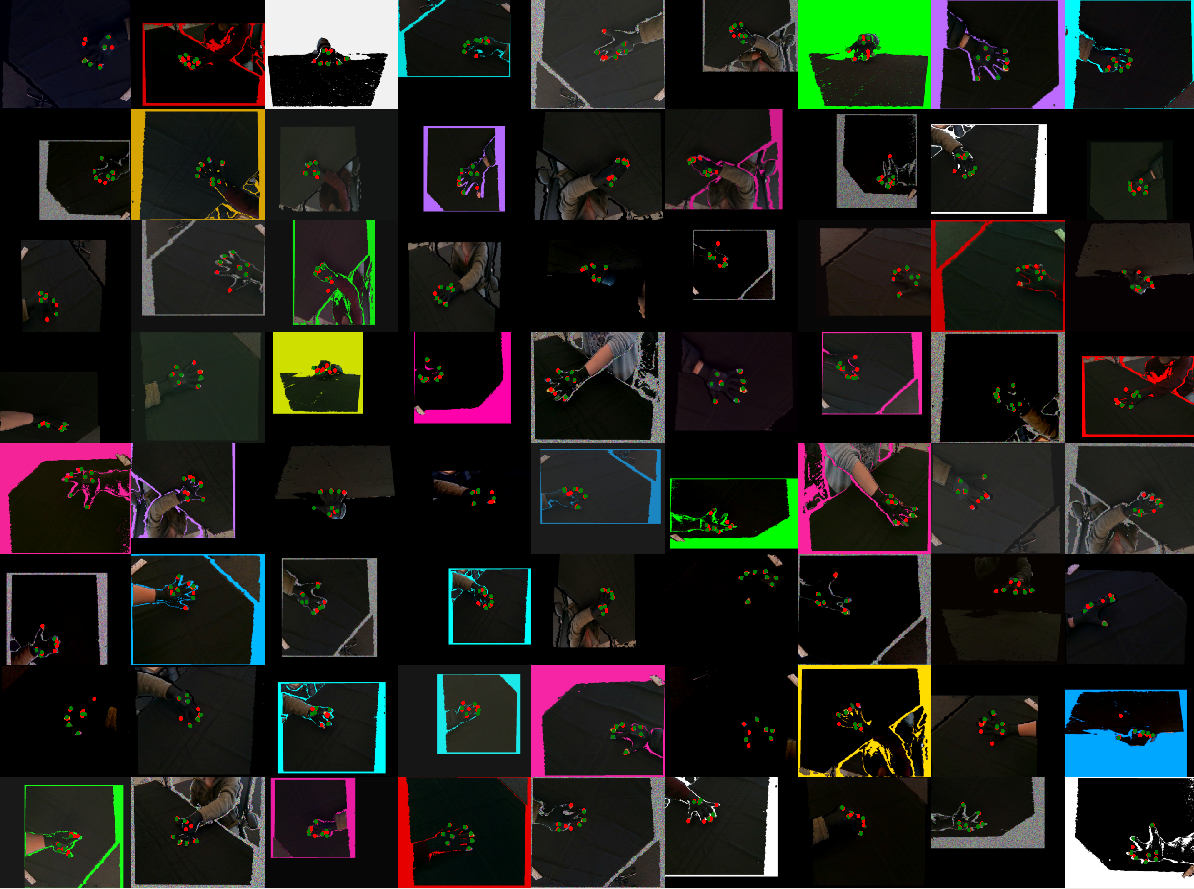}
  \vspace{-5mm}
  \caption{GloveNet is trained with enough data augmentation to allow for accurate and stable keypoint detection of the coloured blobs. The green points represent the predicted while the red keypoints represent the ground truth keypoints. This snapshot is taken while the network is still training. The network is trained on images with annotations equal to or more than 4. Some ground truth annotations are missing due to occlusions or failing depth consistency check across the 4 views.}
  \label{fig:glovenet}
\end{figure*}

\paragraph{Data Augmentation}
We use \texttt{imgaug} \cite{imgaug} and apply various data augmentations while training. Additionally, because we want to focus on the hand moving on the table, for each training image we set the colour values of all pixels with depth beyond a threshold to zero. At training time, we either fill these zeroed out values with any of the colours on the glove or leave the image unchanged based on random number generator. We also replace these zeroed out values random noise based on some probability. All this ensures that the network learns to ignore the colours in the background that look similar to the colours on the glove.

\paragraph{Confidence Estimation of Predictions}
We also obtain confidence for each predicted finger-tip location using test-time augmentation (TTA). We generate new images by shifting the original image by random shifts and pass them all through the network in one batch. We then subtract the applied random shifts from the predicted positions for each image to bring them into the reference frame of the original image and average them out to obtain the mean and standard deviation. We use the standard deviation as our confidence measure. We also use this to clean up the ground truth data that is noisy.

\paragraph{Outlier Rejection}
At test time, we generate four randomly shifted images per camera image and a combined total of 16 images from all four cameras. We compute the predicted finger-tip locations and their confidence measures and discard those that have low confidence. Of the confident ones we compute the euclidean distances $\mathsf{d}_i$ between them and the previous finger-tip locations and turn them into probabilities $\mathsf{p}_i$ via softmax: 

\begin{equation*}
\mathsf{p}_i = \frac{\exp(-\alpha (\mathsf{d}_i - \min_i \mathsf{d}_i))}{\sum_{i=0}^{\mathsf{N}} \exp(-\alpha (\mathsf{d}_i - \min_i \mathsf{d}_i))}    
\end{equation*}

We then push the predicted locations that have probability $\mathsf{p}_i>0.2$ in a rolling buffer and compute the geometric median \cite{Vardi:etal:NAS2000} to obtain the final predicted location of the finger-tip in 3D. The hyper-parameter $\alpha = 500$.

\newcommand{\blocka}[2]{\multirow{3}{*}{\(\left[\begin{array}{c}\text{3$\times$3, #1}\\[-.1em] \text{3$\times$3, #1} \end{array}\right]\)$\times$#2}
}
\newcommand{\blockb}[3]{\multirow{3}{*}{\(\left[\begin{array}{c}\text{1$\times$1, #2}\\[-.1em] \text{3$\times$3, #2}\\[-.1em] \text{1$\times$1, #1}\end{array}\right]\)$\times$#3}
}

\begin{table*}[h]
\begin{center}
\resizebox{0.6\linewidth}{!}{
%\footnotesize
\begin{tabular}{c|c|c}
\hline
layer name & output size & parameters\\
\hline
\multirow{1}{*}{input} &  \multirow{1}{*}{320$\times$240}  &  - \\
\hline
\multirow{2}{*}{conv1} &  \multirow{2}{*}{160$\times$120}  & 7$\times$7, 64, stride 2 \\
  &  & 3$\times$3 max pool, stride 2 \\
\hline
\multirow{3}{*}{conv2} &  \multirow{3}{*}{80$\times$60}  & \blockb{256}{64}{3} \\
  &  & \\
  &  & \\
\hline
\multirow{3}{*}{conv3} &  \multirow{3}{*}{40$\times$30}  & \blockb{512}{128}{4} \\
  &  & \\
  &  & \\
\hline
\multirow{1}{*}{conv\_transpose} &  \multirow{1}{*}{80$\times$60}  & 3$\times$3, 8 \\
\hline
\multirow{1}{*}{spatial\_softmax} &  8$\times$2  & $\beta$=50 \\
\hline
\end{tabular}
}
\end{center}
\caption{Architectures for GloveNet. Downsampling in conv2 and conv3 is performed with a stride of 2. We regress to 8 keypoint locations --- 5 keypoints for fingers and 3 on the back of the palm to obtain the hand pose. We scale the predicted keypoint locations by a factor of 4 to obtain the results for 320$\times$240 resolution image. The pre-trained weights come from anti-aliased version of ResNet-50 as done in \cite{Zhang:etal:ICML19}. The $\beta$ is softmax temperature.
}
\label{tab:arch}
\vspace{-2em}
\end{table*}

We found that while the predictions of the blobs at the back of the palm were stable, the predictions of finger-tips blobs tended to be quite inconsistent across time. Since the annotations were generated by computing the center of mass (CoM) of the segmented blob using the HSV colour thresholding in OpenCV, the CoM of the finger-tip were somewhat inconsistent across frames due to occlusions. Therefore, we relied only on the hand pose estimate provided by the blobs at the back of the palm.

\subsection{Architecture for Hand Pose Estimation with PointNet++}
The PointNet++ implementation we used in this paper is from \url{https://github.com/sshaoshuai/Pointnet2.PyTorch}.

\begin{table}[h]
\centering
\resizebox{0.6\linewidth}{!}{
\begin{tabular}{c|c|c|c}
\hline
layer name & mlp features & radius & num points\\
\hline
SA$_{1}$     & [3, 64, 64, 128]  & 0.2 & 2048 \\
%\hline
SA$_{2}$     & [128, 128, 128, 256] & 0.4 & 1024 \\
%\hline
SA$_{3}$     & [256, 128, 128, 256] & 0.8 & 512 \\ 
%\hline
SA$_{4}$     & [256, 128, 128, 256] & 1.2 & 256 \\
\hline
FP$_{4}$     & [256+256, 256, 256] &  &  512 \\ 
%\hline
FP$_{3}$     & [256+256, 256, 256] &  &  1024 \\ 
%\hline
FP$_{2}$     & [256+128, 256, 256] &  &  2048 \\
%\hline
FP$_{1}$     & [256+3, 256, 256] &  &  8192\\ 
\hline
\end{tabular}
}
\caption{The architecture is composed of 4 set abstraction layers, SA$_{i}$ and 4 feature propagation layers, FP$_{j}$. The set abstraction layer sub-samples the points while the feature propagation layer interpolates features at a higher resolution.
}
\end{table}

\paragraph{Set Abstraction} A set abstraction level takes $N \times (d + C)$ as input of $N$ points with $d$-dim coordinates and $C$-dim point feature. It outputs tensor of $N' \times (d + C')$ where $N'$ sub-sampled points with $d$-dim coordinates and new $C'$-dim feature vectors summarise local context. 

\paragraph{Feature Propagation} In a feature propagation level, point features are propagated from $N_i \times (d + C)$ points to $N_{i-1}$ points where $N_{i-1}$ and $N_i$ (with $N_i \leq N_{i-1}$) are point set size of input and output of set abstraction level $i$. It is achieved by interpolating feature values of $N_i$ points at coordinates of the $N_{i-1}$ points. The interpolated features on $N_{i-1}$ points are then concatenated with skip linked point features from the set abstraction level.

\paragraph{Predicting Keypoint Locations} The backbone of the hand pose estimation is PointNet++ architecture which returns an embedding, $f$, of size $N \times C$. Different MLPs are used to map this embedding to the corresponding desired outputs.

\begin{align*}
z & =  \texttt{mlp\_layer1}(f) \\
\delta_{xyz} & = \texttt{voting}(z) \\
\texttt{coords} & = \texttt{input}_{xyz} + \delta_{xyz} \\ 
\\ 
\texttt{JointMask}_{xyz} & = \texttt{sigmoid}(\texttt{seg}(z)) \\ 
\texttt{HandSeg}_{xyz} & = \texttt{cls}(z) \\ 
\texttt{HandSegProb}_{xyz} & = \texttt{sigmoid}(\texttt{HandSeg}_{xyz}) \\ 
\\ 
\texttt{weights} & = \texttt{HandSegProb}_{xyz} \cdot \texttt{JointMask}_{xyz} \\
\texttt{Keypoints} & = \frac{\sum \texttt{weights} \cdot \texttt{coords}}{\sum \texttt{weights}}
\label{eq:1}
\end{align*}

\begin{table}[h]
\centering
\resizebox{0.35\linewidth}{!}{
\begin{tabular}{c|c}
\hline
layer name & parameters \\
\hline
\texttt{mlp\_layer1}     & [256, 256, 256]  \\
%\hline
\texttt{voting}     & [256, 23$\times$3] \\
%\hline
\texttt{seg}     & [256, 23]  \\ 
%\hline
\texttt{cls}    & [256, 2] \\ 
\hline
\end{tabular}
}
\caption{Various MLPs used to map embedding to the corresponding outputs.}
\end{table}

The \texttt{voting} layer obtains the relative positions, $\delta_{xyz}$, of the 23 keypoints with respect to each point. The \texttt{seg} layer obtains the masks for each keypoint \textit{i.e.} the neighbourhood of points that contribute to the location of a keypoint. The \texttt{HandSeg} layer segments hand from the background. We use Euclidean losses for both \texttt{voting} as well as \texttt{Keypoints} while a sigmoid cross-entropy is used for \texttt{HandSeg}.

\subsection{Architecture for JointNet}
The 23$\times$3 keypoint locations are unrolled to a 69-dimensional vector before feeding to the JointNet which returns a 20-dimensional vector of joint angles. Of all the hand-designed architectures we tried, we found this particular architecture to be an optimal trade-off between accuracy and efficiency.
\begin{table}[h]
\centering
\resizebox{0.3\linewidth}{!}{
\begin{tabular}{c|c}
\hline
layer name & parameters \\
\hline
linear1     & 69$\times$128  \\
%\hline
linear2     & 128$\times$256 \\
%\hline
linear3     & 256$\times$20  \\ 
\hline
\end{tabular}
}
\caption{The JointNet architecture is comprised of three layers. The layers linear1 and linear2 also use BatchNorm1d and ReLU.
}
\end{table}

\subsection{Completion Times Over 5 Consecutive Trials}
We show the completion times for the 5 consecutive trails for each of the tasks. The failed trial is denoted by \textbf{F}. Note that the for most of the trials, the pilot only used 3-4 training trails to warm up. These 5 consecutive trails allow for testing both the ability to carry out a certain task without getting tired as well as showcasing that the tracking works without failures. Admittedly, the performace can vary depending on the pilot and how they are feeling on a given day but our experiments have revealed that the performance is in general quite similar. 

\begin{center}
\begin{tabular}{|p{5.0cm}|>{\centering}p{2.0cm}|p{1.0cm}|p{1.0cm}|p{1.0cm}|p{1.0cm}|p{1.0cm}|c|c|}
\hline
Task                                   & Pilots  & \multicolumn{5}{l|}{Completion Times for 5 Consecutive Trials(s)} & Mean & Std. \\ \hline
 \rowcolor{LightCyan} & Pilot 1 &     19    &  16       &  17       &     11   &    18    &  16     &  3.11     \\ \cline{2-9} 
\rowcolor{LightCyan}\multirow{1}{*}{\texttt{Pick and Place: Brick}} & Pilot 2 &    22     &     22   &    19    &   16     &    14    &  19    &  3.57     \\ \hline
 & Pilot 1 &     28    &  14       &  15       &     16   &    23    &  19     &  6.05     \\ \cline{2-9} 
\multirow{1}{*}{\texttt{Pick and Place: Spam}}  & Pilot 2 &    23     &     23   &    28    &   29     &    20    &  25    &  3.78     \\ \hline             
\rowcolor{LightCyan} & Pilot 1 &     27    &  26       &  32       &     38   &    35    &  32     &  5.12     \\ \cline{2-9} 
\rowcolor{LightCyan}\multirow{1}{*}{\texttt{Card Sliding}}   & Pilot 2 &    18     &     12   &    18    &   15     &    17    &  16    &  2.54     \\ \hline  
                                      
& Pilot 1 &     50    &  18       &  20       &     29   &    18    &  27     &  13.6     \\ \cline{2-9} 
\multirow{1}{*}{\texttt{Pick and Place: Pringles}}   & Pilot 2 &    25     &     53   &    29    &   36     &    63    &  41    &  16.22     \\ \hline                 
  \rowcolor{LightCyan} & Pilot 1 &     48     &     67   &   \cellcolor{Fail}\textbf{F}    &  \cellcolor{Fail} \textbf{F}     &    58    &  58    &  9.50     \\ \cline{2-9} 
  \rowcolor{LightCyan} \multirow{1}{*}{\texttt{Brick Gaiting}} & Pilot 2 &    37     &     44   &   \cellcolor{Fail} \textbf{F}    &  \cellcolor{Fail} \textbf{F}     &    28    &  36    &  8.02     \\ \hline
& Pilot 1 &     38    &  42       &  32       &    \cellcolor{Fail} \textbf{F}   &    28    &  35     &  6.21     \\ \cline{2-9} 
\multirow{1}{*}{\texttt{Pouring}}   & Pilot 2 &    73     &     56   &    62    &   50     &    57    &  60    &  8.61     \\ \hline
  \rowcolor{LightCyan}  & Pilot 1 &     51    &  39       &  45       &     \cellcolor{Fail} \textbf{F}   &    77    &  53     &  16.73     \\ \cline{2-9} 
  \rowcolor{LightCyan} \multirow{1}{*}{\texttt{Box Flip}}   & Pilot 2 &    174     &     26   &    90    &   30     &    67    &  77    &  60.18     \\ \hline
& Pilot 1 &     41    &  49       &  54       &     45   &    165    &  71     &  52.87     \\ \cline{2-9} 
\multirow{1}{*}{\texttt{Blocks (L)}}   & Pilot 2 &    53     &     93   &    79    &   43     &    61    &  66    &  20.12     \\ \hline
  \rowcolor{LightCyan} & Pilot 1 &     89    &  66       &  79       &     77   &    75    &  77     &  8.25     \\ \cline{2-9} 
  \rowcolor{LightCyan}\multirow{1}{*}{\texttt{Peanut Jar}}                                     & Pilot 2 &    68     &     105   &    84    &   87     &    57    &  80    &  18.45     \\ \hline
 & Pilot 1 &     64    &  94       &  70       &    \cellcolor{Fail} \textbf{F}   &    71    &  75     &  13.2     \\ \cline{2-9} 
\multirow{1}{*}{\texttt{Cup Insertion}}   & Pilot 2 &    125     &     \cellcolor{Fail} \textbf{F}   &    124    &   124     &    112    &  121    &  6.18     \\ \hline

 \rowcolor{LightCyan}  & Pilot 1 &     48    &  115       &  170       &     58   &    154    &  109     &  55.00     \\ \cline{2-9} 
 \rowcolor{LightCyan} \multirow{1}{*}{\texttt{Tea}} & Pilot 2 &    54     &     48   &    99    &   105     &    213    &  104    &  66.22     \\ \hline
 & Pilot 1 &     179    &  278       &  64       &     80   &    298    &  180     &  108.37    \\ \cline{2-9} 
\multirow{1}{*}{\texttt{Blocks (M)}}  & Pilot 2 &    99     &     48   &    82    &   75     &    152    &  91    &  38.63     \\ \hline
\rowcolor{LightCyan}  & Pilot 1 &     105    &  66       &  195       &     96   &    63    &  105     &  61.82     \\ \cline{2-9} 
 \rowcolor{LightCyan} \multirow{1}{*}{\texttt{Wallet}}                                     & Pilot 2 &    321     &     92   &    328    &   100     &    218    &  212    &  114.36     \\ \hline
& Pilot 1 &     136    &  371       &  169       &     \cellcolor{Fail}\textbf{F}   &    484    &  290     &  165.88     \\ \cline{2-9} 
\multirow{1}{*}{\texttt{Blocks (S)}}                           & Pilot 2 &    113     &     89   &    69    &   117     &    67    &  91    &  23.57     \\ \hline
\rowcolor{LightCyan}  & Pilot 1 &     442    &  271       &  375       &     297   &    405    &  358     &  72.18     \\ \cline{2-9} 
\rowcolor{LightCyan}\multirow{1}{*} {\texttt{Container}} & Pilot 2 &    189     &     212   &    258    &   238     &    243    &  228    &  27.39    \\ \hline
\end{tabular}
\end{center}

\subsection{Retargeting With Neural Networks}
We also tried retargeting with neural networks but found the results to be unsatisfactory --- it did not provide the accuracy commensurate with the online optimisation with sequential least squares. Moreover, the projection threshold used in retargeting can require some tuning when grasping small objects and therefore it becomes cumbersome to train a neural network for new arbitrary task.

\subsection{Model-based and Model-free Hand Tracking}

\begin{figure*}[htp]
  \centering
  \includegraphics[width=\linewidth]{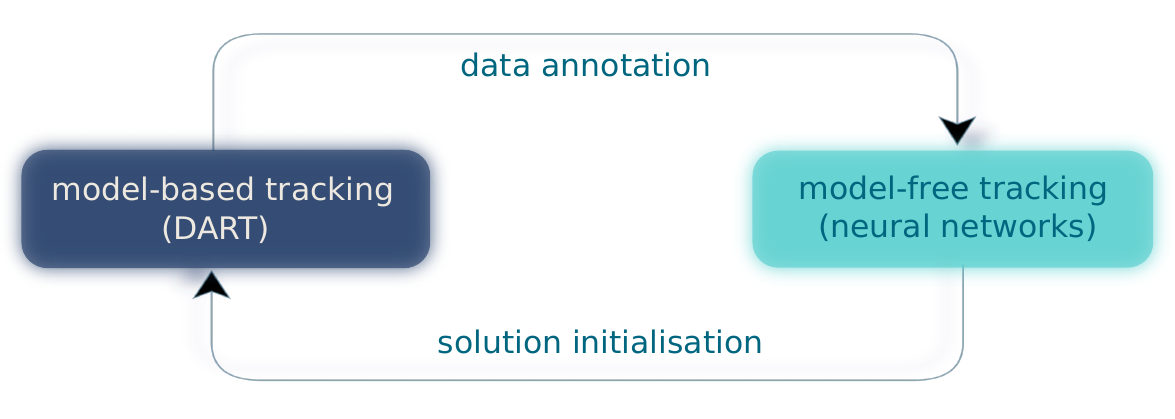}
  \vspace{-5mm}
  \caption{The combination of model-based and model-free can improve hand tracking significantly enabling long duration tracking without any failures.}
  \label{fig:mbmf}
\end{figure*}

Our hand tracking system relies on a combination of model-based and model-free tracking. Model-based tracking systems tend to be more accurate as they optimise online on the input observations given the model. However, since the optimisation tends to be highly non-linear, they also need a good initialisation to find a sensible solution. This motivates us to use a model-free system which can provide good initialisation. Our model-free system is a neural network trained on the data generated by model-based system.

We use model-based tracker in DART \cite{Schmidt:etal:RSS2014} and collect data in the regions where it works reliably and do this repeatedly to cover a wide range of poses. The performance of DART can be stochastic: it may work for the same motion reliably at times and fail catastrophically at other times due to spurious local minima in the optimisation given the input point cloud. However, if we collect data for the scenarios where it works reasonably well, we can use a neural network to fit this data and ensure that it can provide good initialisation for DART preventing it from falling into the spurious local minima in future. This is incumbent on the fact that neural networks can generalise slightly outside the range of training set --- this happens to be true in our case here. We can do this procedure of data collection and neural network fitting repeatedly and improve the performance of DART such that tracking works without any failures for long duration. Our two stage PointNet++ based architecture is trained on the annotations generated by DART and allows us to make the tracking both robust and accurate by providing good initialisation.

\subsection{Software Tools}
Different software tools used in this paper are described below in the table.
\begin{center}
\begin{tabular}{|>{\centering}p{4.0cm}|>{\centering}p{5.0cm}|p{7.0cm}|}
\hline
\multicolumn{1}{|c|}{\textbf{Software Tool}} & \textbf{Purpose}                                                                                 & \multicolumn{1}{c|}{\textbf{Source}}                                                                                                        \\ \hline
Pangolin                                     & \begin{tabular}[c]{@{}c@{}}Real-time 3D visualisation \\ and plotting\end{tabular}               & \begin{tabular}[c]{@{}l@{}}C++: https://github.com/stevenlovegrove/Pangolin\\ Python binding: https://github.com/uoip/pangolin\end{tabular} \\ \hline
\rowcolor{LightCyan} zmq                                          & \multicolumn{1}{l|}{Lightweight Publisher / Subscriber}                                          & Python: https://zeromq.org/                                                                                                                 \\ \hline
ROS                                          & \multicolumn{1}{l|}{Message passing and Visualisation}                                           & Python: http://wiki.ros.org/rospy                                                                                                           \\ \hline
\rowcolor{LightCyan} NLopt                                        & \begin{tabular}[c]{@{}c@{}}Non-linear Optimisation used in \\ Kinematic Retargeting\end{tabular} & C++: https://nlopt.readthedocs.io/en/latest/                                                                                                \\ \hline
Tensorflow                                   & Training neural networks                                                                         & Python: https://www.tensorflow.org/                                                                                                         \\ \hline
\rowcolor{LightCyan} Pytorch                                      & Training neural networks                                                                         & Python: https://pytorch.org/                                                                                                                \\ \hline
\multicolumn{1}{|c|}{DART}    & Real-time model based \\  hand tracking                                                                          & C++: https://github.com/tschmidt23/dart                                                                                                     \\ \hline
\end{tabular}
\end{center}

\twocolumn

\end{document}